\documentclass[sigconf]{acmart}
\AtBeginDocument{%
  }

\copyrightyear{2026}
\acmYear{2026}
\setcopyright{cc}
\setcctype{by}
\acmConference[MM '26] {Proceedings of the 34th ACM International Conference on Multimedia}{November 10--14, 2026}{Rio de Janeiro, Brazil}
\acmBooktitle{Proceedings of the 34th ACM International Conference on Multimedia (MM '26), November 10--14, 2026, Rio de Janeiro, Brazil}
\acmISBN{979-8-4007-2213-4/2026/11}
\acmDOI{10.1145/3767308.3836039}

\usepackage{multirow}
\usepackage{enumitem}
\newcommand{\scorestd}[2]{#1{\tiny$\pm$#2}}

\usepackage{balance}   % Balances the two columns on the last page (Sheridan requirement).

\usepackage{tabularx}
\usepackage{pifont}
\newcommand{\cmark}{\textcolor{green!60!black}{\ding{51}}}
\newcommand{\xmark}{\textcolor{red!70!black}{\ding{55}}}
\newcommand{\pmark}{\textcolor{orange!80!black}{$\boldsymbol{\sim}$}}
\usepackage{algorithm}
\usepackage{algorithmic}
\usepackage{listings}
\lstdefinestyle{promptbox}{
  basicstyle=\ttfamily\footnotesize,
  breaklines=true,
  breakindent=0pt,
  breakatwhitespace=false,
  postbreak=\mbox{\textcolor{gray}{$\hookrightarrow$}\space},
  frame=single,
  rulecolor=\color{black!30},
  backgroundcolor=\color{black!3},
  xleftmargin=3pt,
  xrightmargin=3pt,
  aboveskip=6pt,
  belowskip=6pt,
  columns=fullflexible,
  keepspaces=true,
  literate={-}{-}1,
}

\begin{document}

\title{MM-ContextFold: Context Folding for Multimodal Agentic Retrieval}

%% Author information must match the ACM rights form exactly.
\author{Yang Tian}
% \authornote{Work done during an internship at Kuaishou.}
\orcid{0009-0003-8559-0600}
\affiliation{%
  \institution{\normalsize Shandong University}
  \city{Jinan}
  \country{China}}
\email{tianyangchn@gmail.com}

\author{Fan Liu}
\authornote{Corresponding authors.}
\orcid{0000-0002-4547-3982}
\affiliation{%
  \institution{\normalsize Southeast University}
  \city{Nanjing}
  \country{China}}
\email{liufancs@gmail.com}

\author{Jingyuan Zhang}
\orcid{0000-0001-6644-4673}
\affiliation{%
  \institution{\normalsize Kuaishou}
  \city{Beijing}
  \country{China}}
\email{zhangjingyuan1994@gmail.com}

\author{Zhenyang Li}
\orcid{0000-0002-4694-1231}
\affiliation{%
  \institution{\normalsize Hong Kong University of Science and Technology}
  \city{Hong Kong}
  \country{Hong Kong}}
\email{zhenyanglidz@gmail.com}

\author{Yupeng Hu}
\orcid{0000-0002-5653-8286}
\affiliation{%
  \institution{\normalsize Shandong University}
  \city{Jinan}
  \country{China}}
\email{huyupeng@sdu.edu.cn}

\author{Liqiang Nie}
\authornotemark[1]
\orcid{0000-0003-1476-0273}
\affiliation{%
  \institution{\normalsize Harbin Institute of Technology (Shenzhen)}
  \city{Shenzhen}
  \country{China}}
\email{nieliqiang@gmail.com}

\renewcommand{\shortauthors}{Yang Tian et al.}

\begin{abstract}
Multimodal Agentic Retrieval (MAR) requires agents to solve complex information-seeking tasks by iteratively invoking external tools. Typical frameworks such as ReAct maintain raw multimodal inputs and the accumulating interaction history in a single, ever-growing context, leading to the context explosion problem. While existing methods alleviate this issue by compressing redundant text, effective strategies for managing token-intensive visual content remain largely underexplored. To address this gap, we first conduct a systematic empirical study of approximately 10,000 trajectories. The results show that as visual cues are progressively extracted through external tools and textualized into the context, raw images become increasingly redundant. Continued image retention is associated with higher output entropy and can even degrade task accuracy. Motivated by these findings, we propose MM-ContextFold, a training-free framework that loads raw images only when needed. It maintains a persistent, text-only main context for high-level planning and spawns ephemeral branch contexts for image-dependent subtasks. Within each branch, the agent loads the relevant images, completes the subtask, and folds the result back into the main context as a concise textual summary; the images and branch trace are then discarded. Experiments on seven MAR benchmarks across five backbone models show that MM-ContextFold improves average accuracy by 6.3 percentage points over ReAct while reducing the working context length by 27.5\%.

\end{abstract}





\maketitle
%% acmart/hyperref leave the PDF Author field empty, so write it into the Info dictionary directly.
\pdfinfo{/Author (Yang Tian, Fan Liu, Jingyuan Zhang, Zhenyang Li, Yupeng Hu, Liqiang Nie)}

\section{Introduction}
\label{sec:intro}
Multimodal Agentic Retrieval (MAR)~\cite{geng2026webwatcher,zhou2026worldvqa,zeng2026vision,tao2026mmsearch} has recently emerged as a challenging task setting in deep research~\cite{team2025tongyi,team2025mirothinker,chu2026redsearcher,li2025websailor}. Such tasks typically involve identifying long-tail entities and scenarios depicted in query images, as well as gathering time-sensitive or fine-grained information about them. These demands often lie beyond the reach of a model's parametric knowledge or a single retrieval step. MAR agents therefore build on reasoning frameworks such as ReAct~\cite{yao2023react,chu2026redsearcher,zeng2026vision,chen2025mindwatcher} to plan and iteratively invoke external tools, retaining the multimodal input while appending each step's reasoning text, tool calls, and observations to a single, ever-growing context. In long-horizon MAR tasks, this unconstrained accumulation of multimodal information rapidly leads to the context explosion problem, where redundant retrieval results and trial-and-error traces overwhelm the model's reasoning capacity~\cite{ye2026agentfold, sun2026scaling, wu2025resum}.

\begin{figure}[t]
\centering
\includegraphics[width=0.99\linewidth]{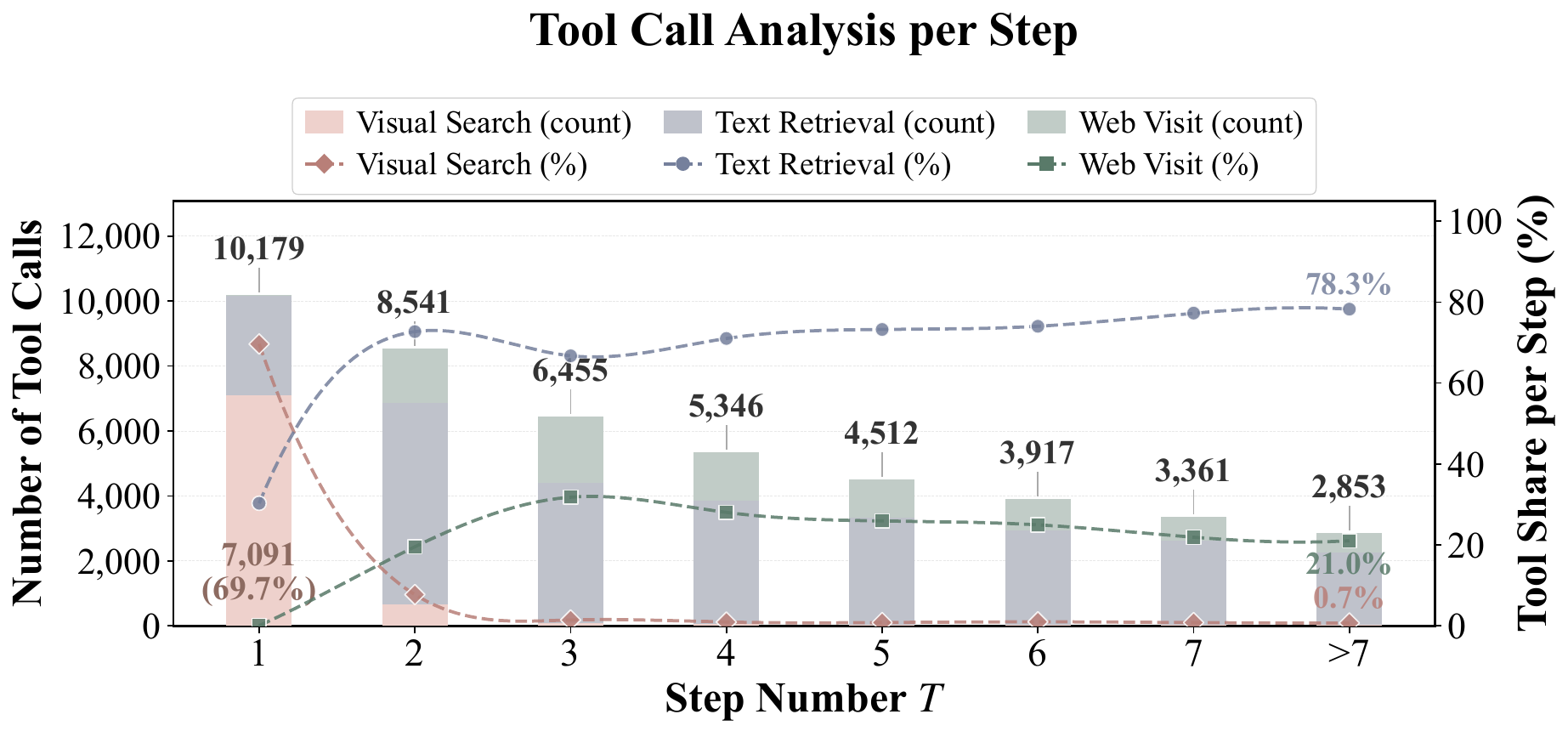}
\vspace{-0.2cm}
\caption{Per-step tool-call counts (bars) and proportions (lines) for \textit{Visual Search}, \textit{Text Retrieval}, and \textit{Web Visit}.}
\Description{Bar chart with overlaid line plots showing, for each trajectory step, the number and share of \textit{Visual Search}, \textit{Text Retrieval}, and \textit{Web Visit} tool calls. \textit{Visual Search} dominates at step 1 with 69.7 percent of calls and declines sharply afterwards, while \textit{Text Retrieval} and \textit{Web Visit} dominate from step 2 onward.}
\vspace{-0.4cm}
\label{fig:intro}
\end{figure}

We further investigate this problem by examining which tools introduce external evidence at successive retrieval steps. Specifically, we collect approximately 10,000 trajectories and analyze the per-step use of three external tools: \textit{Visual Search}, \textit{Text Retrieval}, and \textit{Web Visit}. As shown in Figure~\ref{fig:intro}, evidence acquisition exhibits a stage-wise pattern. \textit{Visual Search} calls, through which the agent grounds the visual content of the query images, are concentrated in the early steps; thereafter, evidence acquisition is dominated by \textit{Text Retrieval} and \textit{Web Visit}. While the resulting redundant textual observations can be compressed by existing methods~\cite{wu2025resum,ye2026agentfold, sun2026scaling}, no comparable mechanism governs the retention of raw images: these token-intensive inputs persist at every reasoning step even when no longer directly needed, continuously exacerbating the context explosion problem~\cite{xu2026rethinking}. This naturally raises the question: \textit{are raw images still necessary throughout the entire retrieval trajectory?}

\begin{figure*}[t]
\centering
\includegraphics[width=0.99\textwidth]{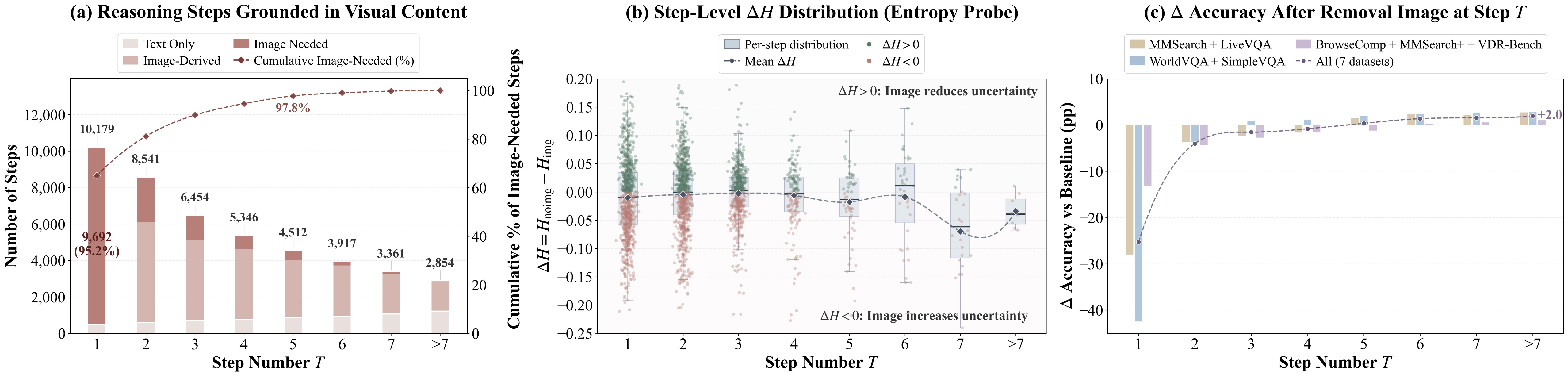}
\vspace{-0.2cm}
\caption{\textbf{Empirical analysis on the role of images in MAR trajectories.} \textbf{(a)}~Per-step breakdown of reasoning texts into \textit{Image Needed} (grounded in new visual information directly extracted from the raw images), \textit{Image-Derived} (based on visual information already textualized in earlier steps), and \textit{Text Only}; the dashed line shows the cumulative proportion of \textit{Image Needed} instances. \textbf{(b)}~Step-level entropy change induced by image removal, measured as $\Delta H_t = \mathbf{H}(\tau_t^{\text{IR}}) - \mathbf{H}(\tau_t)$; box plots show the per-step distributions and the dashed line connects the step-level means. \textbf{(c)}~$\Delta$Accuracy (\%) relative to the with-image baseline when images are removed from step~$T$ onward; grouped bars show the three benchmark categories, and the dashed line shows the weighted average over all seven benchmarks.}
\Description{Three-panel figure. Panel (a): per-step proportions of reasoning texts labeled Image Needed, Image-Derived, and Text Only, with a dashed cumulative Image Needed curve that saturates within the first five steps. Panel (b): box plots of the entropy change caused by image removal at each step; the mean change is slightly negative at steps 1 and 2, stays near zero during steps 3 to 6, and shifts to more negative values from step 7 onward. Panel (c): grouped bars of accuracy change when images are removed from step T onward for three benchmark categories, with a dashed weighted-average line rising from negative values to slightly positive as T increases.}
\vspace{-0.3cm}
\label{fig:pilot}
\end{figure*}

To answer this question, we conduct a systematic empirical study that goes beyond tool-call statistics (\S\ref{subsec:empirical_analysis}). Specifically, we classify each reasoning step by whether it draws directly on raw images, relies on previously textualized visual evidence, or is independent of the images; probe, via output entropy, whether images still implicitly influence the model's output distribution; and measure how removing images from a given step onward affects task accuracy (Figure~\ref{fig:pilot}). Together, these analyses show that the agent's reliance on raw images is concentrated in an early grounding phase, during which task-relevant visual cues are progressively textualized into the context. Beyond this phase, retaining images increases output uncertainty and can degrade task accuracy~\cite{tian2026more,ghosal2026visref}. However, the duration of this phase varies with each task's visual complexity, making any fixed removal step unlikely to generalize across tasks.

Motivated by these findings, we propose \textbf{MM-ContextFold}, a training-free context management framework that enables on-demand loading and timely release of raw images within the agent's working context. MM-ContextFold introduces a dual-state mechanism that separates high-level planning from visual grounding: the agent alternates between a text-only \emph{main state} and an ephemeral multimodal \emph{branch state}. In the main state, the agent reasons over a persistent main context that maintains task progress, the overall strategy, and distilled evidence. When new visual cues are required, the agent initiates an image-dependent subtask and transitions to the branch state, where it loads the relevant raw images into a scoped branch context and executes the subtask through interleaved reasoning and tool calls. Upon completing the subtask, the agent folds its findings back into the main context as a concise textual summary; the raw images and the local branch trace are then discarded. By confining raw images to bounded branch scopes, MM-ContextFold allows textualized evidence to continuously enrich the main context, enabling more informed subsequent planning, while avoiding the accumulation of visual tokens. Experiments on seven MAR benchmarks across five backbone models show that MM-ContextFold improves average accuracy by 6.3 percentage points over the standard ReAct baseline while reducing working context length by 27.5\%. Code and data are released at \url{https://github.com/iLearn-Lab/MM26-MMContextFold}. In summary, our main contributions are as follows:

\begin{itemize}[leftmargin=*,topsep=0pt,partopsep=0pt]
    \item We conduct an in-depth empirical study of approximately 10,000 multimodal retrieval trajectories, providing, to our knowledge, the first systematic empirical characterization of the dynamically diminishing pattern of agent reliance on images in MAR, and demonstrating that retaining images after visual cues have been textualized can degrade reasoning accuracy.
    \item Based on this insight, we propose MM-ContextFold, a dual-state context architecture that structurally decouples visual-grounding steps from the main reasoning trajectory, achieving on-demand loading and timely release of raw images.
    \item Extensive experiments on seven MAR benchmarks across five backbone models show that MM-ContextFold reduces working context length by 27.5\% while improving average accuracy by 6.3 percentage points over the standard ReAct baseline.
\end{itemize}

\section{Related Work}
\textbf{Multimodal Agentic Retrieval.} LLM-based agents have demonstrated strong capabilities in autonomous information seeking, evolving from text-only deep research systems~\cite{li2025webthinker,wu2025webwalker,wu2025webdancer,zheng2025deepresearcher} to multimodal agentic retrieval (MAR), where queries involve images alongside text~\cite{yang2026multimodal,geng2026webwatcher,zhou2026worldvqa,zeng2026vision,tao2026mmsearch}. Since neither the model's parametric knowledge nor a single retrieval step suffices for such queries, MAR requires agents to iteratively invoke external search tools to gather, verify, and integrate multimodal evidence. This iterative process distinguishes MAR from multimodal RAG pipelines, which operate over closed or pre-collected corpora without agentic tool interaction~\cite{zhang2026mldocrag,wang2026vimrag}. To evaluate MAR, several benchmarks have been proposed~\cite{jiang2025mmsearch,fu2025seeking,zeng2026vision,deng2026deepimagesearch}. On the system side, the ReAct framework~\cite{yao2023react}, already a mainstream architecture for text-only deep research agents, has been further adopted as the backbone of MAR systems. Recent efforts~\cite{geng2026webwatcher,zhang2026vsearcher,chng2025sensenova} mainly focus on improving agent capabilities through reinforcement learning (RL). While these training-centric efforts advance MAR, they retain the append-only context mechanism inherited from ReAct. Under this mechanism, the context explosion problem observed in text-only agents~\cite{wu2025resum,ye2026agentfold} is further amplified by the persistent accumulation of token-intensive visual content alongside textual interaction traces.

\textbf{Context Management for Search Agents.}
Context management for LLM agents broadly falls into two categories: \emph{external context augmentation}, which injects knowledge from outside the current trajectory, such as user profiles or past conversations~\cite{li2025memos,xu2025mem,chhikara2025mem0}, and \emph{intra-task context curation}, which manages the context generated within the task itself over long horizons~\cite{mei2025survey,qiao2025webresearcher}. Early intra-task methods adopt stepwise summarization that compresses the full history at each turn~\cite{zhou2026mem1,yu2026memagent,wu2025resum}, but such uniform compression risks discarding fine-grained details critical for complex tasks. Two representative works move beyond this limitation by introducing \emph{agentic context folding}. AgentFold~\cite{ye2026agentfold} treats context as a dynamic cognitive workspace, enabling the agent to selectively fold multi-step interactions at different scales, from condensing individual steps to consolidating completed sub-investigations, and maintains only 7K tokens after 100 interaction turns. ContextFold~\cite{sun2026scaling} proposes a branch-and-return mechanism with FoldGRPO, an end-to-end RL algorithm that trains the agent's branching behavior through dense process rewards to achieve high context compression. Despite this progress, existing methods mainly target text-only scenarios. When extended to multimodal retrieval, they face a structural mismatch: raw images are needed only \emph{locally} along the trajectory, and their grounding results can be preserved as compact text, yet uniformly compressing or persistently retaining the images either loses fine-grained visual details or inflates the context and degrades accuracy. This mismatch motivates our work.

\section{MM-ContextFold} 
\label{sec:pilot}
This section first formalizes the MAR task and its standard agentic framework (\S\ref{subsec:mar-task}), then presents an empirical study on the role of raw images across retrieval trajectories (\S\ref{subsec:empirical_analysis}), whose findings motivate the design of the MM-ContextFold framework (\S\ref{sec:method}).

\subsection{Task Formulation}
\label{subsec:mar-task}
We formalize MAR as a sequential decision-making process. Given a multimodal input $(q,\mathcal{I})$, where $q$ denotes a natural-language question and $\mathcal{I}=\{I_k\}_{k=1}^K$ denotes a set of query images, the agent $\mathcal{M}_\theta$ produces an answer $a$ through multi-step interaction with an external environment. At each step $t$, conditioned on the accumulated context $\tau_t$, the agent generates a reasoning text $r_t$ and invokes a tool $u_t \in \mathcal{U}$; the returned observation $o_t$, together with $(r_t, u_t)$, is then appended to the context to form $\tau_{t+1}$:
\begin{equation}
\left\{
\begin{aligned}
&\;\tau_t := (q,\mathcal{I},r_1, u_1, o_1, \dots, r_{t-1}, u_{t-1}, o_{t-1}), \\
&\;(r_t, u_t) \leftarrow \mathcal{M}_\theta(\tau_t).
\end{aligned}
\right.
\label{eq:react}
\end{equation}
In the setting examined in this paper, the tool set $\mathcal{U}$ comprises three tool types:  \textit{Visual Search}, which retrieves visual references and source links based on $\mathcal{I}$; \textit{Text Retrieval}, which performs open-domain retrieval and returns relevant webpage metadata and links; and \textit{Web Visit}, which accesses a specified URL and extracts the parsed content of the webpage.

\subsection{Empirical Analysis}
\label{subsec:empirical_analysis}
As shown in Figure~\ref{fig:intro}, the tool-call distribution of approximately 10,000 ReAct trajectories\footnote{Cross-model trajectories span Gemini-3-Flash, GPT-5.2, Qwen3.5-9B, Qwen3.5-27B, and Qwen3.5-35B-A3B. Trajectories are collected via both API services and local deployment; of the roughly 10,000 collected trajectories, 5,087 valid ones are from locally deployed Qwen-series models. Because token-level probabilities are available only from local deployment, we conduct the entropy probe on this locally deployed Qwen subset; for consistency, the paired step-level image-removal ablations are also run on the same subset.} reveals that \textit{Visual Search} is concentrated in the first two steps, after which the agent shifts almost exclusively to \textit{Text Retrieval} and \textit{Web Visit}. This behavioral pattern provides preliminary evidence that the agent's reliance on raw images is confined to the early stages. However, tool-call distributions characterize visual reliance only in the action component $u_t$; they cannot reveal whether the reasoning component $r_t$ remains grounded in raw image content, whether images implicitly influence the model's output distribution, or how their removal affects final task accuracy. We therefore analyze the role of raw images from three complementary angles: the visual grounding of reasoning texts (\S\ref{subsec:pilot-surface-deep}), a stepwise entropy probe on output distributions (\S\ref{subsec:pilot-entropy}), and step-level image-removal ablations on task accuracy (\S\ref{subsec:pilot-acc}).

\subsubsection{Visual Grounding of Reasoning Text}
\label{subsec:pilot-surface-deep}
To characterize how the reasoning text $r_t$ at each step is grounded in the visual input, we employ Gemini-3-Flash as an annotator and label every reasoning step in the collected trajectories.\footnote{We validate annotation quality on a random subsample of 500 steps independently labeled by three human annotators; the LLM annotations achieve a 93.5\% agreement rate with the majority human vote.} For each step $t$, the annotator receives the preceding context $\tau_t$ defined in Eq.~(\ref{eq:react}) together with $r_t$, which it assigns to one of three mutually exclusive categories: (1) \textit{Image Needed}, where $r_t$ introduces visual information drawn directly from the query images $\mathcal{I}$ that has not appeared earlier in $\tau_t$; (2) \textit{Image-Derived}, where $r_t$ relies on visual information already textualized in earlier steps rather than on the raw images themselves; and (3) \textit{Text Only}, where $r_t$ draws on neither source of visual information. Figure~\ref{fig:pilot}(a) shows that 95.2\% of Step-1 reasoning texts are labeled \textit{Image Needed}, but this share falls sharply thereafter, and \textit{Image-Derived} becomes the dominant category from Step~2 onward, while \textit{Text Only} rises steadily. Overall, 97.8\% of all steps labeled \textit{Image Needed} occur within the first five steps. Together with the tool-call distribution in Figure~\ref{fig:intro}, these results indicate that the agent's direct use of raw images is restricted to an early visual-grounding phase, during which task-relevant visual information is extracted and converted into textual artifacts.

\subsubsection{Entropy Probe}
\label{subsec:pilot-entropy} 
The analysis of output text $(r_t, u_t)$ shows that raw images are not uniformly necessary throughout the retrieval trajectory. However, this conclusion does not rule out the possibility that raw images still implicitly influence the model's output distribution at later steps. To test this, we introduce a stepwise output-entropy probe~\cite{kadavath2022language, kuhn2023semantic, li2026rethinking}. Specifically, let $\tau_t$ denote the original multimodal context at step $t$ as defined in Eq.~(\ref{eq:react}), and $\tau_t^{\text{IR}}$ denote the context obtained by removing the query images $\mathcal{I}$ from $\tau_t$. For the reasoning text $r_t = (y_1^{(t)}, \dots, y_{m_t}^{(t)})$, where $y_i^{(t)}$ denotes the $i$-th token and $m_t$ the sequence length, its length-normalized entropy under context $\tau_t$ is:
\begin{equation}
\mathbf{H} (\tau_t)
=
-\frac{1}{m_t}
\sum_{i=1}^{m_t}
\sum_{v\in\mathcal{V}}
p\!\left(v \mid \tau_t, y_{<i}^{(t)}\right)
\log p\!\left(v \mid \tau_t, y_{<i}^{(t)}\right),
\end{equation}
where $\mathcal{V}$ denotes the vocabulary and $p(v|\tau_t, y_{<i}^{(t)})$ is the next-token probability assigned by the model $\mathcal{M}_\theta$. We further define the entropy change induced by image removal:
\begin{equation}
\Delta H_t = \mathbf{H} (\tau_t^{\text{IR}}) - \mathbf{H} (\tau_t).
\end{equation}
Negative $\Delta H_t$ indicates that retaining images increases output entropy. As shown in Figure~\ref{fig:pilot}(b), $\Delta H_t$ exhibits a clear stepwise pattern: at Steps~1--2, $\Delta H_t$ is slightly negative, suggesting that without visual input the model falls back on language priors~\cite{lee2025vlind, wang2026vgr, zhou2025proreason}, yielding more confident yet less visually informed predictions; during Steps~3--6, $|\Delta H_t|$ remains near zero, suggesting that the output distribution is largely unaffected by image removal; and from Step~7 onward, $\Delta H_t$ shifts to more negative values, indicating that beyond the visual-grounding steps, raw images may introduce redundant context into the predominantly textual reasoning~\cite{li2025redundancylens, sun2025mitigating}.

\subsubsection{Image-Removal Ablation}
\label{subsec:pilot-acc}
We further conduct ablations using $\tau_t^{\text{IR}}$, i.e., the context with query images removed, to measure the direct impact on task accuracy. Starting from step $t$, the model continues the standard ReAct loop based on $\tau_t^{\text{IR}}$ until termination. Let $T_i$ denote the total number of steps in the original trajectory of sample $i$, and $\mathcal{X}_t = \{i \mid T_i \ge t\}$ the set of samples whose original trajectories reach step $t$. Let $\hat{a}_i^{\text{IR},t}$ and $\hat{a}_i$ denote the final answers of the image-removed and original runs, respectively. We report the accuracy difference over $\mathcal{X}_t$:
\begin{equation}
\Delta\mathrm{Acc}(t) = \frac{1}{|\mathcal{X}_t|} \sum_{i \in \mathcal{X}_t} \left( \mathbf{1}\!\left[\hat{a}_i^{\text{IR},t} = a_i^{*}\right]-\mathbf{1}\!\left[\hat{a}_i = a_i^{*}\right]\right).
\end{equation}
As shown in Figure~\ref{fig:pilot}(c), removing images during the first two steps causes the most severe accuracy degradation ($-25.2$\% at Step~1; $-3.9$\% at Step~2), aligning with the output-text analysis in \S\ref{subsec:pilot-surface-deep} where \textit{Visual Search} calls and raw-image-grounded reasoning are both concentrated. As the trajectory progresses, the impact of image removal diminishes steadily: the accuracy change approaches near zero by Step~4, and from Step~6 onward removing images consistently improves accuracy by $+1$\% to $+2$\% on average. We further disaggregate these results by three benchmark categories (\S\ref{subsec:exp-setup}). A clear contrast emerges at Step~5: whereas \textit{Atomic factuality} (WorldVQA \cite{zhou2026worldvqa} and SimpleVQA \cite{cheng2025simplevqa}) and \textit{Dynamic information-seeking} benchmarks (MMSearch \cite{jiang2025mmsearch} and LiveVQA \cite{fu2025seeking}) already yield positive gains, \textit{Visual deep research} benchmarks (BrowseComp-VL \cite{geng2026webwatcher}, MMSearch-Plus \cite{tao2026mmsearch}, and VDR-Bench \cite{zeng2026vision}), which involve more complex queries and longer retrieval trajectories, still show a slightly negative $\Delta\mathrm{Acc}$. This divergence across benchmarks suggests that although removing images in later steps can improve accuracy, the optimal removal step varies by task, making it difficult to define a single removal point that generalizes across benchmarks.

\subsubsection{Key Findings}
\label{subsec:pilot-summary}
The three analyses above converge on a picture of how raw images contribute to MAR trajectories: (1) the agent's reliance on raw images is concentrated in specific visual-grounding steps; once the task-relevant visual cues are extracted and textualized, subsequent reasoning operates predominantly over textual artifacts (Figure~\ref{fig:intro}, Figure~\ref{fig:pilot}(a)); (2) beyond the grounding phase, persistently retaining raw images provides little benefit (Figure~\ref{fig:pilot}(c)); instead, it introduces redundant visual context that increases output entropy, suggesting higher reasoning uncertainty (Figure~\ref{fig:pilot}(b)); and (3) the number of steps for which raw images remain beneficial varies across benchmark categories: \textit{Atomic factuality} and \textit{Dynamic information-seeking} tasks complete visual grounding earlier, whereas complex \textit{Visual deep research} tasks require sustained image access over more steps (Figure~\ref{fig:pilot}(c)).

\begin{figure*}[t]
\centering
\includegraphics[width=0.96\linewidth]{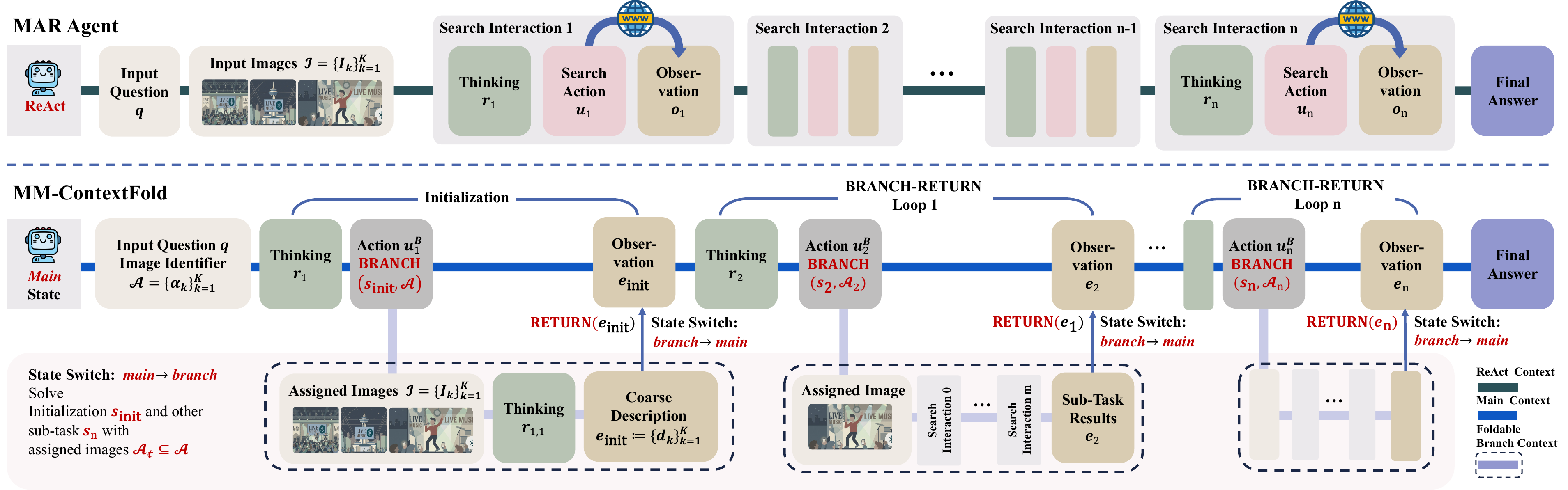}
\vspace{-0.2cm}
\caption{\textbf{Overview of MM-ContextFold.} \textit{Top}: the standard MAR framework, where raw query images $\mathcal{I}$ persist in the context throughout the entire trajectory. \textit{Bottom}: MM-ContextFold maintains a text-only \textit{main state} that plans subtasks using coarse descriptions from initialization; the agent injects only the assigned images into an ephemeral \textit{branch state} for visual grounding and tool use, and \texttt{RETURN} folds the branch result back as a textual summary, after which the branch context is discarded.}
\Description{Two-panel schematic. Top: the standard MAR framework keeps raw query images in a single growing context at every reasoning step. Bottom: MM-ContextFold keeps a text-only main state for planning, spawns an ephemeral branch state that loads only the assigned images for visual grounding and tool calls, and folds the branch result back into the main state as a textual summary before the branch context is discarded.}
\vspace{-0.2cm}
\label{fig:overview}
\end{figure*}

\subsection{MM-ContextFold Framework}
\label{sec:method}
The empirical findings motivate three design principles: (P1) \textit{decoupled reasoning and grounding}, separating textual planning and evidence synthesis from visual grounding to reflect their distinct cognitive roles; (P2) \textit{ephemeral visual context}, confining raw images to scoped contexts rather than retaining them persistently, since their presence can complicate later text-dominant reasoning; and (P3) \textit{adaptive image allocation}, determining when and which images to load to accommodate varying visual demands across tasks. To instantiate these principles, MM-ContextFold introduces a dual-state mechanism (Figure~\ref{fig:overview}): the agent alternates between a text-only \textbf{main state} for planning and an ephemeral multimodal \textbf{branch state} for scoped grounding (P1); each branch loads only the designated images (P3), and upon completion, its findings are folded back as text and the branch context is discarded (P2).

\subsubsection{Dual-State Mechanism}
\label{subsec:unified-interface}
Given a natural language question $q$ and a set of query images $\mathcal{I}=\{I_k\}_{k=1}^{K}$, we first assign each image a stable textual identifier $\alpha_k$ and construct an image registry:
\begin{equation}
\mathcal{P} = \{(\alpha_k, I_k)\}_{k=1}^{K}.
\end{equation}
In the main state, the agent $\mathcal{M}_\theta$ starts from the initial main context $\tau_1 := (q, \mathcal{A})$, where $\mathcal{A}=\{\alpha_k\}_{k=1}^{K}$ is the set of image identifiers. The transitions between the two states are governed by two actions:
\begin{itemize}[leftmargin=*]
    \item $u_t^B=\texttt{BRANCH}(s_t, \mathcal{A}_t)$: Initiates a subtask $s_t$ that requires visual information from a subset of images $\mathcal{A}_t \subseteq \mathcal{A}$ (possibly empty), triggering a transition to the branch state.
    \item $u_t^R=\texttt{RETURN}(e_t)$: Concludes the current subtask. The agent folds the gathered multimodal evidence into a structured textual summary $e_t$ and returns to the main state.
\end{itemize}
A complete cycle consists of the agent generating a reasoning step $r_t$ and a \texttt{BRANCH} action in the main state, executing the subtask in the branch state, and concluding with a \texttt{RETURN} action. Subsequently, the ephemeral branch trajectory is discarded, and the persistent main context $\tau_t$ is updated with the synthesized result:
\begin{equation}
\label{eq:main-update}
\tau_{t+1} = (\tau_t,\; r_t,\; u_t^B,\; e_t).
\end{equation}
The main and branch policies share the same backbone model $\mathcal{M}_\theta$ but operate under different action spaces: the main state emits only high-level control actions (\texttt{BRANCH} or \texttt{FINISH}), whereas a branch state either invokes a tool from $\mathcal{U}$ or terminates with \texttt{RETURN}.

\subsubsection{Bootstrapping Image-Aware Planning}
\label{subsec:phase1-init}
To support \textbf{adaptive image allocation (P3)}, the agent needs to dynamically decide which images are relevant for subsequent subtasks. However, effective planning requires a basic semantic understanding of the visual inputs that raw identifiers alone cannot provide. We therefore introduce an initialization cycle ($t=1$) before the regular \texttt{BRANCH}--\texttt{RETURN} loop. From the initial main context $\tau_1$, the agent first issues $u_1^B = \texttt{BRANCH}(s_{\mathrm{init}}, \mathcal{A})$:
\begin{equation}
(r_1, u_1^B) \leftarrow \mathcal{M}_\theta(\tau_1),
\end{equation}
where $s_{\mathrm{init}}$ requests high-level overviews of the visual inputs. Upon completing $s_{\mathrm{init}}$, the agent invokes $\hat{u}_{1,1}^R=\texttt{RETURN}(e_{\mathrm{init}})$:
\begin{equation}
(\hat{r}_{1,1}, \hat{u}_{1,1}^R) \leftarrow \mathcal{M}_\theta(\hat{\tau}_{1,1}), \quad \hat{\tau}_{1,1} := (\tau_1, r_1, u_1^B, \mathcal{A}, \mathcal{I}).
\end{equation}
The returned result $e_{\mathrm{init}}:= \{d_k\}_{k=1}^{K}$ consists of a textual description $d_k$ for each image $I_k$. These descriptions are intentionally concise, capturing only high-confidence visual content to serve as reliable planning priors rather than detailed substitutes for raw images. Folding $e_{\mathrm{init}}$ back into the main state via Eq.~(\ref{eq:main-update}) yields:
\begin{equation}
\tau_2 = (\tau_1,\; r_1,\; u_1^B,\; e_{\mathrm{init}}).
\end{equation}

\subsubsection{Iterative Multimodal Branch-and-Return Reasoning}
\label{subsec:phase2-grounding}
Equipped with $\tau_2$, the agent proceeds to the regular iterative loop ($t \ge 2$). Following the principle of decoupled reasoning and grounding (P1), the main state operates over textual artifacts. This separation allows the agent to focus on global task planning and information synthesis without reprocessing raw pixels. Given the current main context $\tau_t$, the agent evaluates whether the accumulated information suffices to answer $q$. If so, it outputs the final answer; otherwise, it plans a new subtask $s_t$ targeting the current information gap:
\begin{equation}
(r_t, u_t) \leftarrow \mathcal{M}_\theta(\tau_t), \quad
u_t \in \bigl\{\texttt{FINISH}(a),\;\; \texttt{BRANCH}(s_t, \mathcal{A}_t)\bigr\}.
\end{equation}
Here, $\mathcal{A}_t \subseteq \mathcal{A}$ dictates the specific images to be injected into the subsequent branch. $\mathcal{A}_t$ can be empty ($\varnothing$) if the required visual grounding has already been completed in prior iterations, enabling on-demand visual access. For a \texttt{BRANCH} action, the branch context is initialized by inheriting the main context $\tau_t$ alongside the on-demand raw images $\mathcal{I}_t$:
\begin{equation}
\hat{\tau}_{t,1} := \bigl(\tau_t, r_t, u_t^B, \mathcal{A}_t, \mathcal{I}_t\bigr), \quad
\mathcal{I}_t = \{I \mid (\alpha, I) \in \mathcal{P}, \alpha \in \mathcal{A}_t\}.
\end{equation}
This allows the agent to leverage global task background and prior folded results. Inside the branch, the same backbone model rolls out a local policy over the branch context:
\begin{equation}
\left(\hat{r}_{t,j}, \hat{u}_{t,j}\right) \leftarrow \mathcal{M}_\theta(\hat{\tau}_{t,j}), \quad
\hat{u}_{t,j} \in \mathcal{U} \cup \{\texttt{RETURN}(e)\}.
\label{eq:branch-policy}
\end{equation}
If $\hat{u}_{t,j} \in \mathcal{U}$, the branch executes the selected tool, receives observation $\hat{o}_{t,j}$, and appends $(\hat{r}_{t,j}, \hat{u}_{t,j}, \hat{o}_{t,j})$ to the local context to form $\hat{\tau}_{t,j+1}$. If $\hat{u}_{t,j}=\texttt{RETURN}(e_t)$, the branch terminates and sends the textual summary $e_t$ back to the main state. The local trajectory is
\begin{equation}
\hat{\tau}_t := \bigl(\hat{\tau}_{t,1}, \hat{r}_{t,1},\hat{u}_{t,1},\hat{o}_{t,1},\;\dots,\;\hat{r}_{t,J_t},\hat{u}_{t,J_t}\bigr), \quad J_t \le B_{\max},
\end{equation}
where the final action $\hat{u}_{t,J_t}$ is always \texttt{RETURN}$(e_t)$. If the step budget is exhausted before an explicit \texttt{RETURN}, the final branch call is constrained to summarize the partial branch trajectory and terminate. The main context is subsequently updated according to Eq.~(\ref{eq:main-update}). If the subtask is not fully resolved within the current cycle, the updated main context seamlessly carries over the partial progress, allowing the agent to allocate the same or different images in subsequent iterations to continue the grounding process.

\subsubsection{Discussion}
MM-ContextFold can be viewed as a dynamic context management framework that rethinks how visual information is integrated into long-horizon agentic reasoning. Rather than persistently retaining raw images or replacing them entirely with coarse textual surrogates, it operationalizes the three design principles to resolve the tension between reasoning efficiency and perceptual fidelity. The framework preserves direct access to original visual evidence within temporary branch contexts, folding the extracted findings back into the persistent textual trajectory. Unlike in generic context folding, the persistent and temporary states are modality-asymmetric rather than simply two scopes of the same mixed-modality history: the main state remains text-only, while original visual evidence is preserved only inside temporary branch contexts and folded back as text after use.

\section{Experiments}
\label{sec:experiments}

\begin{table*}[t]
    \caption{\textbf{Main results on seven multimodal agentic retrieval benchmarks (accuracy, \%).} Subscripts give the number of evaluated instances per benchmark, identical for all methods and trials. All methods use the same tool set $\mathcal{U}$ and the same inference budget of up to 20 LLM calls. Within each backbone block, the best results are \textbf{bold} and the second-best \underline{underlined}. Avg.\ is weighted by these counts. Each entry reports mean $\pm$ standard deviation over three independent trials.}
    \vspace{-0.2cm}
    \centering
    \scriptsize
    \resizebox{0.97\textwidth}{!}{
    \begin{tabular}{l|ccccccc|c}
    \toprule
    \textbf{Method} & \textbf{BC-VL}\,$_{\#300}$ & \textbf{LiveVQA}\,$_{\#245}$ & \textbf{MMSearch}\,$_{\#171}$ & \textbf{MMS-Plus}\,$_{\#311}$ & \textbf{SimpleVQA}\,$_{\#300}$ & \textbf{VDR-Bench}\,$_{\#200}$ & \textbf{WorldVQA}\,$_{\#200}$ & \textbf{Avg.} \\
    \midrule
    Gemini-3-Flash & \scorestd{44.3}{0.8} & \scorestd{73.5}{0.7} & \scorestd{59.5}{0.9} & \scorestd{22.5}{0.8} & \scorestd{75.0}{0.6} & \scorestd{14.5}{0.7} & \scorestd{40.5}{0.8} & 47.5 \\
    ReAct & \scorestd{\underline{54.7}}{1.6} & \scorestd{79.2}{1.4} & \scorestd{74.1}{1.4} & \scorestd{\underline{37.3}}{1.6} & \scorestd{78.0}{1.1} & \scorestd{\underline{23.0}}{1.4} & \scorestd{57.0}{1.6} & 57.6 \\
    AgentFold & \scorestd{50.3}{1.4} & \scorestd{82.0}{1.1} & \scorestd{\underline{77.6}}{1.2} & \scorestd{30.7}{1.5} & \scorestd{73.7}{1.1} & \scorestd{22.0}{1.1} & \scorestd{56.0}{1.4} & 55.4 \\
    ContextFold & \scorestd{52.3}{1.4} & \scorestd{\underline{82.9}}{1.1} & \scorestd{76.6}{1.1} & \scorestd{34.6}{1.4} & \scorestd{\underline{80.3}}{0.9} & \scorestd{21.0}{1.1} & \scorestd{\underline{57.5}}{1.4} & \underline{57.7} \\
    \textbf{Ours} & \scorestd{\textbf{62.7}}{1.1} & \scorestd{\textbf{85.4}}{0.9} & \scorestd{\textbf{82.8}}{1.1} & \scorestd{\textbf{40.3}}{1.4} & \scorestd{\textbf{82.7}}{0.9} & \scorestd{\textbf{28.5}}{1.2} & \scorestd{\textbf{61.0}}{1.1} & \textbf{63.2} \\
    \midrule
    GPT-5.2 & \scorestd{47.0}{0.7} & \scorestd{61.6}{0.9} & \scorestd{46.8}{0.8} & \scorestd{12.6}{0.7} & \scorestd{67.3}{0.8} & \scorestd{12.5}{0.6} & \scorestd{35.5}{0.9} & 41.1 \\
    ReAct & \scorestd{\underline{55.7}}{1.4} & \scorestd{74.6}{1.6} & \scorestd{69.8}{1.5} & \scorestd{\underline{34.7}}{1.4} & \scorestd{78.0}{1.3} & \scorestd{\underline{22.5}}{1.6} & \scorestd{55.5}{1.4} & \underline{56.0} \\
    AgentFold & \scorestd{48.6}{1.2} & \scorestd{75.6}{1.3} & \scorestd{\underline{75.4}}{1.4} & \scorestd{28.0}{1.3} & \scorestd{72.4}{1.0} & \scorestd{19.8}{1.3} & \scorestd{53.7}{1.2} & 52.8 \\
    ContextFold & \scorestd{50.7}{1.3} & \scorestd{\underline{76.3}}{1.2} & \scorestd{74.9}{1.3} & \scorestd{31.9}{1.2} & \scorestd{\underline{79.2}}{1.1} & \scorestd{18.7}{1.3} & \scorestd{\underline{56.2}}{1.2} & 55.2 \\
    \textbf{Ours} & \scorestd{\textbf{61.3}}{1.0} & \scorestd{\textbf{78.1}}{1.1} & \scorestd{\textbf{80.7}}{0.9} & \scorestd{\textbf{40.7}}{1.2} & \scorestd{\textbf{81.3}}{1.1} & \scorestd{\textbf{26.5}}{1.0} & \scorestd{\textbf{59.5}}{1.3} & \textbf{61.1} \\
    \midrule
    Qwen3.5-9B & \scorestd{25.3}{0.9} & \scorestd{50.7}{0.6} & \scorestd{19.3}{0.7} & \scorestd{2.7}{0.5} & \scorestd{54.3}{0.8} & \scorestd{4.0}{0.6} & \scorestd{20.0}{0.7} & 26.2 \\
    ReAct & \scorestd{41.8}{1.5} & \scorestd{66.5}{1.2} & \scorestd{62.4}{1.4} & \scorestd{19.2}{1.5} & \scorestd{67.8}{1.2} & \scorestd{14.0}{1.2} & \scorestd{44.5}{1.6} & 44.9 \\
    AgentFold & \scorestd{42.3}{1.3} & \scorestd{69.0}{1.2} & \scorestd{64.3}{1.5} & \scorestd{18.2}{1.4} & \scorestd{65.7}{1.3} & \scorestd{13.5}{1.0} & \scorestd{39.0}{1.2} & 44.3 \\
    ContextFold & \scorestd{\underline{44.7}}{1.4} & \scorestd{\underline{71.2}}{1.0} & \scorestd{\underline{65.1}}{1.2} & \scorestd{\underline{21.1}}{1.3} & \scorestd{\underline{69.7}}{1.0} & \scorestd{\underline{17.0}}{1.2} & \scorestd{\underline{46.0}}{1.2} & \underline{47.5} \\
    \textbf{Ours} & \scorestd{\textbf{54.3}}{1.2} & \scorestd{\textbf{76.7}}{1.1} & \scorestd{\textbf{71.3}}{1.0} & \scorestd{\textbf{26.7}}{1.2} & \scorestd{\textbf{74.7}}{1.1} & \scorestd{\textbf{21.5}}{1.1} & \scorestd{\textbf{49.5}}{0.9} & \textbf{53.4} \\
    \midrule
    Qwen3.5-27B & \scorestd{28.7}{0.7} & \scorestd{60.0}{0.9} & \scorestd{25.7}{0.7} & \scorestd{4.7}{0.7} & \scorestd{61.3}{0.6} & \scorestd{5.0}{0.6} & \scorestd{21.5}{0.8} & 30.6 \\
    ReAct & \scorestd{48.0}{1.4} & \scorestd{78.6}{1.5} & \scorestd{74.3}{1.6} & \scorestd{\underline{33.0}}{1.4} & \scorestd{\underline{76.7}}{1.3} & \scorestd{\underline{23.0}}{1.2} & \scorestd{\underline{52.0}}{1.4} & \underline{54.8} \\
    AgentFold & \scorestd{46.7}{1.5} & \scorestd{77.6}{1.2} & \scorestd{68.8}{1.2} & \scorestd{21.3}{1.4} & \scorestd{70.7}{1.2} & \scorestd{18.5}{1.3} & \scorestd{49.5}{1.2} & 49.9 \\
    ContextFold & \scorestd{\underline{51.7}}{1.2} & \scorestd{\underline{80.3}}{1.3} & \scorestd{\underline{75.8}}{1.2} & \scorestd{28.7}{1.2} & \scorestd{72.7}{1.1} & \scorestd{20.5}{1.0} & \scorestd{50.5}{1.2} & 53.9 \\
    \textbf{Ours} & \scorestd{\textbf{60.3}}{1.0} & \scorestd{\textbf{83.5}}{1.0} & \scorestd{\textbf{76.6}}{0.9} & \scorestd{\textbf{38.3}}{1.2} & \scorestd{\textbf{79.3}}{1.0} & \scorestd{\textbf{27.0}}{1.1} & \scorestd{\textbf{57.0}}{1.3} & \textbf{60.3} \\
    \midrule
    Qwen3.5-35B-A3B & \scorestd{29.3}{0.6} & \scorestd{62.0}{0.7} & \scorestd{26.9}{0.9} & \scorestd{4.3}{0.7} & \scorestd{62.3}{0.6} & \scorestd{8.0}{0.4} & \scorestd{22.5}{0.7} & 31.7 \\
    ReAct & \scorestd{47.3}{1.5} & \scorestd{72.5}{1.6} & \scorestd{71.2}{1.3} & \scorestd{\underline{28.3}}{1.4} & \scorestd{70.7}{1.4} & \scorestd{\underline{20.5}}{1.2} & \scorestd{\underline{50.5}}{1.4} & \underline{51.1} \\
    AgentFold & \scorestd{43.3}{1.2} & \scorestd{70.1}{1.4} & \scorestd{67.4}{1.2} & \scorestd{21.3}{1.1} & \scorestd{68.3}{1.2} & \scorestd{17.0}{1.0} & \scorestd{46.0}{1.3} & 47.2 \\
    ContextFold & \scorestd{\underline{48.6}}{1.2} & \scorestd{\underline{75.8}}{1.3} & \scorestd{\underline{73.5}}{1.3} & \scorestd{24.7}{1.3} & \scorestd{\underline{71.7}}{1.1} & \scorestd{18.5}{1.2} & \scorestd{47.0}{1.1} & 51.0 \\
    \textbf{Ours} & \scorestd{\textbf{57.3}}{0.9} & \scorestd{\textbf{80.5}}{1.1} & \scorestd{\textbf{76.2}}{1.2} & \scorestd{\textbf{36.3}}{1.1} & \scorestd{\textbf{77.7}}{1.0} & \scorestd{\textbf{22.5}}{1.2} & \scorestd{\textbf{56.5}}{0.9} & \textbf{58.1} \\
    \bottomrule
    \end{tabular}
    }
    \label{tab:main-results}
    \vspace{-0.3cm}
\end{table*}

\subsection{Experimental Settings}
\label{subsec:exp-setup}
\textbf{Benchmarks.} We evaluate MM-ContextFold on seven multimodal agentic retrieval benchmarks, grouped into three categories by their requirements for external knowledge retrieval and reasoning complexity. \textit{Atomic factuality}: SimpleVQA \cite{cheng2025simplevqa} and WorldVQA \cite{zhou2026worldvqa} require the agent to recognize entities in images, ranging from common to long-tail cases, and retrieve the corresponding encyclopedic knowledge. \textit{Dynamic information-seeking}: MMSearch \cite{jiang2025mmsearch} and LiveVQA \cite{fu2025seeking} involve images related to recent events and specialized domains, requiring the agent to query the web for up-to-date or rare information. \textit{Visual deep research}: BrowseComp-VL (BC-VL) \cite{geng2026webwatcher}, MMSearch-Plus (MMS-Plus) \cite{tao2026mmsearch}, and VDR-Bench \cite{zeng2026vision} present questions where images provide key clues for identifying ambiguous or obfuscated entities. Solving them requires extracting fine-grained visual cues, conducting multi-step retrieval, and synthesizing cross-modal evidence to infer hard-to-find answers. Notably, MMSearch-Plus further involves multi-image queries where relevant clues are distributed across images. The per-benchmark instance counts are reported in Table~\ref{tab:main-results}: for BC-VL, WorldVQA, and VDR-Bench we evaluate on stratified subsets of the larger public pools that preserve the original difficulty and category structure, while the remaining four benchmarks use their released splits directly. All evaluation files are released in our repository.

\textbf{Baselines.} We compare MM-ContextFold with three types of baselines: \textit{direct inference}, \textit{ReAct}, and \textit{context management methods}. Direct inference answers the multimodal query solely from the model's parametric knowledge, without any external tool use. For \textit{ReAct}, we adopt the standard ReAct framework as implemented in \textbf{WebWatcher} \cite{geng2026webwatcher}, a recent high-performing openly released multimodal deep research agent. For \textit{context management methods}, we compare against two representative baselines, \textbf{AgentFold} \cite{ye2026agentfold} and \textbf{ContextFold} \cite{sun2026scaling}. Both are designed for text-only scenarios, as context management for multimodal search agents remains largely unexplored. For fair comparison, all methods use the same backbone, tool APIs, answer format, and total call budget. For AgentFold, we preserve its original multi-scale textual folding logic and expose the same multimodal query and \textit{Visual Search} interface used by our method; the folded memories remain text-only. For ContextFold, we preserve its original branch-and-return controller and context inheritance rule; the method can call the same \textit{Visual Search} tool, but because it has no explicit image-scoping mechanism, raw query images remain globally visible once placed in the main context and are therefore inherited by spawned branches.

\begin{table*}[t]
\caption{Ablation results (accuracy, \%). Parenthesized values indicate the drop relative to MM-ContextFold on the same backbone.}
\vspace{-0.2cm}
\centering
\scalebox{0.850}{%
\begin{tabular}{l|lllllll|l}
\toprule
\textbf{Config} & \textbf{BC-VL} & \textbf{LiveVQA} & \textbf{MMSearch} & \textbf{MMS-Plus} & \textbf{SimpleVQA} & \textbf{VDR-Bench} & \textbf{WorldVQA} & \textbf{Avg.} \\
\midrule
% Gemini-3-Flash & 44.3 & 73.5 & 59.5 & 22.5 & 75.0 & 14.5 & 40.5 & 47.5 \\
MM-ContextFold (Gemini) & 62.7 & 85.4 & 82.8 & 40.3 & 82.7 & 28.5 & 61.0 & 63.2 \\
$\hookrightarrow$ Image Caption & {49.0\;{\scriptsize($-$13.7)}} & {73.5\;{\scriptsize($-$11.9)}} & {65.5\;{\scriptsize($-$17.3)}} & {23.0\;{\scriptsize($-$17.3)}} & {64.7\;{\scriptsize($-$18.0)}} & {13.0\;{\scriptsize($-$15.5)}} & {39.5\;{\scriptsize($-$21.5)}} & {46.9\;{\scriptsize($-$16.3)}} \\
\hspace{0.4cm}w/o Init & {53.3\;{\scriptsize($-$9.4)}} & {75.6\;{\scriptsize($-$9.8)}}& {76.6\;{\scriptsize($-$6.2)}} & {35.3\;{\scriptsize($-$5.0)}} & {73.3\;{\scriptsize($-$9.4)}} & {22.5\;{\scriptsize($-$6.0)}} & {53.5\;{\scriptsize($-$7.5)}} & {55.5\;{\scriptsize($-$7.7)}} \\
\hspace{0.4cm}w/o Image in Branch & {52.7\;{\scriptsize($-$10.0)}} & {80.4\;{\scriptsize($-$5.0)}} & {76.0\;{\scriptsize($-$6.8)}}& {32.3\;{\scriptsize($-$8.0)}} & {76.7\;{\scriptsize($-$6.0)}} & {24.5\;{\scriptsize($-$4.0)}} & {55.5\;{\scriptsize($-$5.5)}} & {56.5\;{\scriptsize($-$6.7)}} \\
\hspace{0.4cm}w/ Image in Main & {51.7\;{\scriptsize($-$11.0)}} & {78.6\;{\scriptsize($-$6.8)}}& {78.9\;{\scriptsize($-$3.9)}} & {35.0\;{\scriptsize($-$5.3)}}& {78.7\;{\scriptsize($-$4.0)}} & {20.5\;{\scriptsize($-$8.0)}} & {56.5\;{\scriptsize($-$4.5)}} & {56.8\;{\scriptsize($-$6.4)}}\\
\midrule
MM-ContextFold (Qwen3.5-35) & 57.3 & 80.5 & 76.2 & 36.3 & 77.7 & 22.5 & 56.5 & 58.1 \\
$\hookrightarrow$ Image Caption & {40.0\;{\scriptsize($-$17.3)}} & {71.4\;{\scriptsize($-$9.1)}} & {40.4\;{\scriptsize($-$35.8)}} & {15.0\;{\scriptsize($-$21.3)}} & {53.0\;{\scriptsize($-$24.7)}} & {9.0\;{\scriptsize($-$13.5)}} & {23.5\;{\scriptsize($-$33.0)}} & {36.7\;{\scriptsize($-$21.4)}} \\
\hspace{0.4cm}w/o Init & {39.0\;{\scriptsize($-$18.3)}} & {70.5\;{\scriptsize($-$10.0)}} & {63.2\;{\scriptsize($-$13.0)}} & {24.3\;{\scriptsize($-$12.0)}} & {60.7\;{\scriptsize($-$17.0)}} & {15.5\;{\scriptsize($-$7.0)}} & {46.5\;{\scriptsize($-$10.0)}} & {45.1\;{\scriptsize($-$13.0)}} \\
\hspace{0.4cm}w/o Image in Branch & {44.3\;{\scriptsize($-$13.0)}} & {72.1\;{\scriptsize($-$8.4)}} & {65.1\;{\scriptsize($-$11.1)}}& {22.6\;{\scriptsize($-$13.7)}} & {67.7\;{\scriptsize($-$10.0)}} & {19.0\;{\scriptsize($-$3.5)}} & {51.5\;{\scriptsize($-$5.0)}} & {48.4\;{\scriptsize($-$9.7)}} \\
\hspace{0.4cm}w/ Image in Main & {47.6\;{\scriptsize($-$9.7)}} & {76.7\;{\scriptsize($-$3.8)}} & {71.3\;{\scriptsize($-$4.9)}}& {22.7\;{\scriptsize($-$13.6)}} & {72.3\;{\scriptsize($-$5.4)}} & {17.5\;{\scriptsize($-$5.0)}} & {49.0\;{\scriptsize($-$7.5)}} & {50.6\;{\scriptsize($-$7.5)}} \\
\bottomrule
\end{tabular}%
}
\label{tab:ablation}
\vspace{-0.2cm}
\end{table*}

\textbf{Implementation.} We evaluate all methods on five backbones via OpenRouter, namely Gemini-3-Flash, GPT-5.2, and the Qwen3.5 series (9B, 27B, 35B-A3B). All methods share the same tool set $\mathcal{U}$ (\S\ref{subsec:mar-task}): \textit{Text Retrieval} takes a free-form query generated by the agent and returns the top 10 results with titles, snippets, and URLs via the Serper API; \textit{Visual Search} takes an image identifier $\alpha_k$, resolves it to the image $I_k$ via the registry $\mathcal{P}$, and submits $I_k$ to the Google Lens API, which returns the top 10 visually similar results with their captions and webpages; \textit{Web Visit} fetches the parsed content of a model-specified URL via Jina Reader. We cap interaction steps per branch at $B_{\max} = 5$ and total LLM calls at $T_{\max} = 20$; every main- and branch-state generation counts as one LLM call, so all methods operate under the same budget. We report accuracy averaged over three independent trials; run-to-run variation reflects both decoding nondeterminism and the changing responses of the live retrieval APIs. Correctness is determined by a fixed LLM-as-Judge protocol following prior work~\cite{geng2026webwatcher, team2025tongyi}: the same judge model (Gemini-3-Flash) and prompt are used for all methods and backbones, reducing variance from answer-style differences.

\subsection{Main Results}
\label{subsec:main-results}
\textbf{Overall performance.} As shown in Table~\ref{tab:main-results}, MM-ContextFold consistently outperforms all baselines on every backbone. On average across the five backbones, it reaches 59.2\%, compared with 52.9\% for ReAct, 53.1\% for ContextFold, and 49.9\% for AgentFold, corresponding to gains of $+$6.3, $+$6.1, and $+$9.3 percentage points (pp), respectively. The improvements are most pronounced on the three \textit{Visual deep research} benchmarks, where direct inference yields only 18.7\% accuracy on average and the tasks demand longer and more complex interaction trajectories (10.1 tool calls per trajectory on average vs.\ 6.5 on the other four benchmarks). On these three benchmarks, MM-ContextFold outperforms ContextFold, the strongest overall baseline, by 8.2\,pp on average. In contrast, on the remaining four benchmarks, where direct inference already reaches 50.2\% accuracy on average, MM-ContextFold's gains over the three agentic retrieval baselines narrow (e.g., to $+$4.4\,pp over ContextFold).

\textbf{Comparison with context-management baselines.} AgentFold progressively compresses earlier observations into multi-scale textual summaries. While this design improves context efficiency, visual-to-text compression may discard fine-grained details, and successive folding can further reduce the fidelity of summaries derived from earlier visual observations. Consistent with this, AgentFold trails ReAct baseline (49.9\% vs.\ 52.9\% on average). ContextFold provides branch-and-return control, but its inherited state remains modality-symmetric: once raw images enter the main context, they propagate into every subsequent branch regardless of whether the current subtask still requires them. MM-ContextFold instead adopts a modality-asymmetric inheritance pattern by keeping the persistent main state text-only and injecting only the selected raw images into a branch on demand.  This targeted visual access yields gains over ContextFold of $+$5.5 to $+$7.1\,pp across the five backbones.

\begin{table}[t]
\caption{Average accuracy (\%) of fixed-step image-removal heuristics. R@$k$ denotes ReAct-Removal@$k$, which removes the query images from step $k$ onward.}
\vspace{-0.2cm}
\centering
\scalebox{0.85}{%
\begin{tabular}{l|ccccc|cc}
\toprule
\textbf{Backbone} & \textbf{R@1} & \textbf{R@2} & \textbf{R@3} & \textbf{R@4} & \textbf{R@5} & \textbf{ReAct} & \textbf{Ours} \\
\midrule
Gemini & 47.3 & 55.2 & 56.4 & 58.7 & \underline{59.4} & 57.6 & \textbf{63.2} \\
Qwen3.5-35 & 27.6 & 46.4 & 48.8 & 51.7 & \underline{52.4} & 51.1 & \textbf{58.1} \\
\bottomrule
\end{tabular}%
}
\label{tab:removal-heuristic}
\vspace{-0.3cm}
\end{table}

\subsection{Ablation Study}
We conduct ablations on Gemini-3-Flash (Gemini) and Qwen3.5-35B-A3B (Qwen3.5-35) with four variants as shown in Table~\ref{tab:ablation}: \textbf{Image Caption} uses the initialization to textualize query images, after which the agent executes subtasks with only \textit{Text Retrieval} and \textit{Web Visit}; \textbf{w/o Init} skips the initialization cycle (\S\ref{subsec:phase1-init}), leaving the agent in the main state planning only with the image identifiers $\mathcal{A}$; \textbf{w/o Image in Branch} disables raw image allocation to branch contexts, which rely solely on textualized results from initialization; \textbf{w/ Image in Main} injects raw images into the main context, removing the visual scoping constraint. 

\textbf{Ablation of the initialization cycle.} Removing the initialization cycle (w/o Init) reduces accuracy by 7.7\,pp on Gemini and 13.0\,pp on Qwen3.5-35, with pronounced drops on benchmarks such as BrowseComp-VL and LiveVQA. The image descriptions produced during initialization enable the agent in the main state to decompose the task into subtasks. Planning from image identifiers $\mathcal{A}$ alone is not sufficient for these decisions.

\textbf{Ablation of image exclusion in branch contexts.} Both Image Caption and w/o Image in Branch exclude raw images from branch contexts. Image Caption incurs the largest drop among all variants ($-$16.3\,pp on Gemini; $-$21.4\,pp on Qwen3.5-35), as replacing raw images with captions deprives the agent of fine-grained details essential for accurate grounding. In contrast, w/o Image in Branch shows a smaller decline ($-$6.7\,pp on Gemini; $-$9.7\,pp on Qwen3.5-35), as the agent can still invoke \textit{Visual Search} by specifying image identifiers $\alpha_k \in \mathcal{A}$ and can coarsely cross-validate returned metadata against the visual priors from initialization. Nevertheless, without direct pixel access, both variants underperform MM-ContextFold on tasks requiring fine-grained visual matching.

\textbf{Ablation of visual scoping.} The w/ Image in Main variant injects raw images into the main context, removing MM-ContextFold's visual scoping constraint. Average accuracy drops by 6.4\,pp on Gemini and 7.5\,pp on Qwen3.5-35, falling back to nearly the level of ContextFold (Gemini: 56.8\% vs.\ 57.7\%; Qwen3.5-35: 50.6\% vs.\ 51.0\%). The degradation is moderate on \textit{Atomic factuality} and \textit{Dynamic information-seeking} benchmarks (e.g., $-$3.9\,pp on MMSearch with Gemini) but more pronounced on \textit{Visual deep research} tasks such as BrowseComp-VL ($-$11.0\,pp on Gemini; $-$9.7\,pp on Qwen3.5-35), where longer trajectories keep redundant visual tokens in the context for more steps. This aligns with the pilot-study observation (\S\ref{subsec:pilot-acc}) that the marginal contribution of raw images diminishes after the early grounding steps.

\textbf{Comparison with fixed-step image removal.} We evaluate ReAct-Removal@$k$ (R@$k$), which drops the query images from step $k$ onward. As shown in Table~\ref{tab:removal-heuristic}, R@1, where the agent never observes the raw images, severely degrades accuracy relative to ReAct ($-$10.3\,pp on Gemini), underscoring the importance of images during early grounding. Accuracy then improves monotonically with $k$ and surpasses ReAct at $k \ge 4$, indicating that image retention beyond the grounding phase on average hurts accuracy. Yet even the best heuristic (R@5) trails MM-ContextFold by 3.8\,pp on Gemini and 5.7\,pp on Qwen3.5-35. Since the optimal removal point varies across task types (Figure~\ref{fig:pilot}(c)), no single fixed $k$ suits all tasks, motivating adaptive, per-subtask image allocation.

\subsection{Token Usage Analysis}

\begin{table}[t]
\caption{Token usage and tool calls averaged per sample across five backbones and seven benchmarks. Working Context denotes the maximum tokens passed to a single LLM call. Tool calls are broken down by type.}
\centering
\vspace{-0.2cm}
\scalebox{0.85}{%
\begin{tabular}{l|rrrr}
\toprule
\textbf{Metric} & \textbf{ReAct} & \textbf{AgentFold} & \textbf{ContextFold} & \textbf{Ours} \\
\midrule
Cumulative Tokens & 54,551 & 15,993 & 49,594 & 44,088 \\
Working Context & 8,846 & 3,415 & 7,691 & 6,411 \\
\midrule
Tool Calls & 9.6 & 6.8 & 8.7 & 8.1 \\
$\hookrightarrow$ \textit{Text Retrieval} & 7.4 & 4.4 & 5.4 & 5.8 \\
\hspace{0.4cm}\textit{Web Visit} & 1.5 & 1.8 & 2.2 & 1.4 \\
\hspace{0.4cm}\textit{Visual Search} & 0.7 & 0.6 & 1.1 & 0.9 \\
Avg. Branches & - & - & 2.7 & 2.3 \\
\bottomrule
\end{tabular}%
}
\vspace{-0.3cm}
\label{tab:efficiency}
\end{table}

\begin{table}[t]
% \caption{MMSearch-Plus: accuracy(\%) and tool calls (Visual Search / total) by the number of query images. Parenthesized counts denote the number of samples in each group.}
\caption{MMSearch-Plus: accuracy(\%) and tool calls (\textit{Visual Search} / total) by the number of query images, averaged over five backbones. Parenthesized counts denote the number of samples in each group.}
\vspace{-0.2cm}
\centering
\small
\scalebox{0.87}{%
\begin{tabular}{l|cc|cc|cc}
\toprule
\multirow{2}{*}{\textbf{Method}} & \multicolumn{2}{c|}{\textbf{1 Image (222)}} & \multicolumn{2}{c|}{\textbf{2 Images (54)}} & \multicolumn{2}{c}{\textbf{3+ Images (35)}} \\
\cmidrule(lr){2-3} \cmidrule(lr){4-5} \cmidrule(lr){6-7}
& Acc. & Tools & Acc. & Tools & Acc. & Tools \\
\midrule
ReAct & 31.8 & 0.9 / 10.1 & 28.8 & 0.6 / 9.8 & 24.6 & 0.7 / 9.2 \\
AgentFold & 24.9 & 0.7 / 7.3 & 21.4 & 1.2 / 8.2 & 20.9 & 2.2 / 9.2 \\
ContextFold & 30.2 & 0.7 / 5.3 & 24.1 & 2.1 / 12.4 & 22.8 & 2.7 / 16.3 \\
MM-ContextFold & 38.1 & 1.0 / 8.9 & 35.6 & 1.7 / 11.4 & 27.6 & 2.3 / 14.4 \\
\bottomrule
\end{tabular}%
}
\vspace{-0.4cm}
\label{tab:case-mmsplus}
\end{table}

\begin{figure}[t]
\centering
\includegraphics[width=0.99\linewidth]{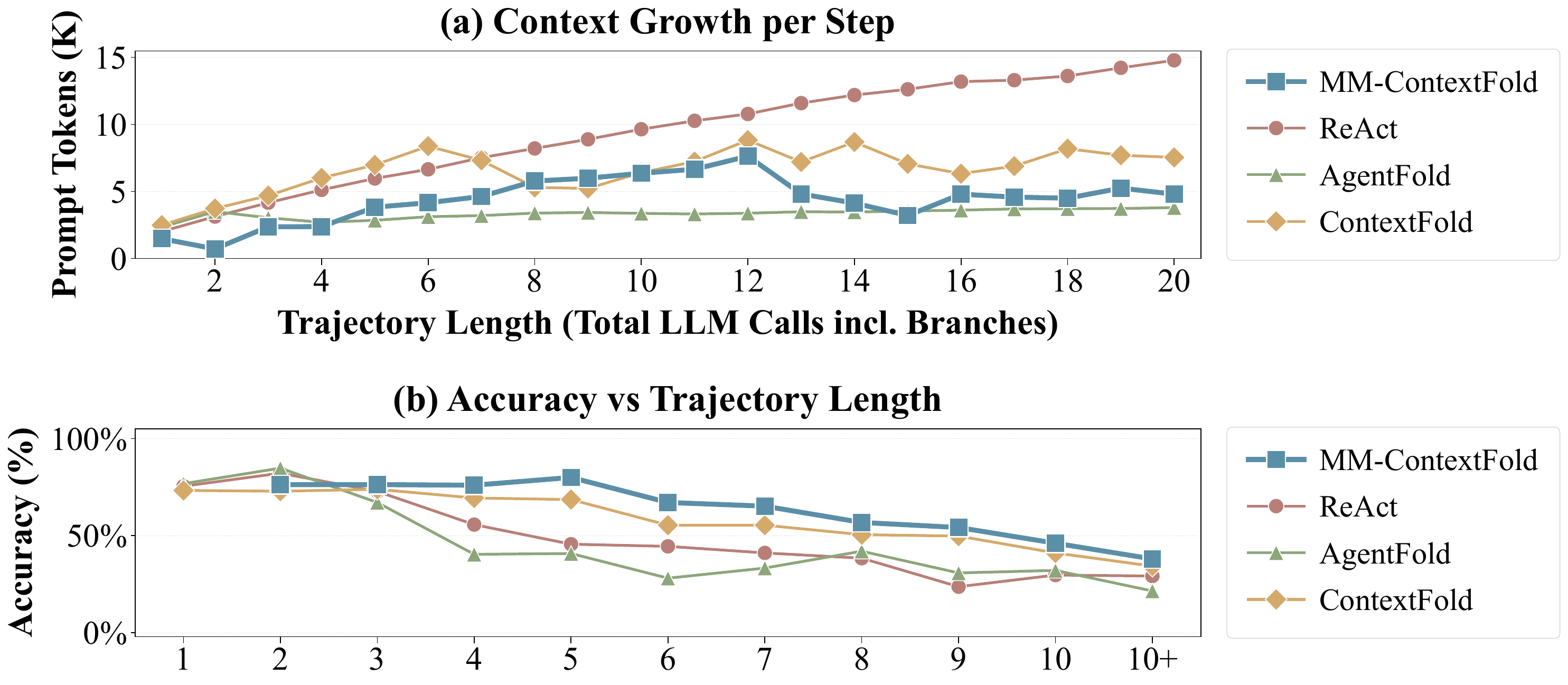}
\vspace{-0.1cm}
\caption{Context growth (a) and accuracy (b) as a function of trajectory length (total LLM calls), averaged across five backbones and seven benchmarks.}
\Description{Two line charts averaged across five backbones and seven benchmarks. Panel (a): working-context tokens versus trajectory length; ReAct grows nearly linearly, ContextFold and MM-ContextFold flatten after early growth, and AgentFold stays lowest. Panel (b): accuracy versus trajectory length; all methods decline on longer trajectories while MM-ContextFold remains highest throughout.}
\vspace{-0.5cm}
\label{fig:acc-token}
\end{figure}

Table~\ref{tab:efficiency} and Figure~\ref{fig:acc-token} compare the four methods on token consumption and on accuracy as a function of trajectory length. Cumulative Tokens denotes the total API token cost per sample, and Working Context the maximum number of tokens passed to a single LLM call. AgentFold is the most economical on both metrics (16.0K / 3.4K), followed by MM-ContextFold (44.1K / 6.4K), which reduces the working context length by 27.5\% relative to ReAct (54.6K / 8.8K), with ContextFold (49.6K / 7.7K) in between. Tool-call counts are comparable across methods, with \textit{Visual Search} accounting for 0.6 to 1.1 calls. These token metrics, however, do not fully capture deployment cost: since MM-ContextFold introduces an initialization phase and branch-orchestration overhead, its lower working context reflects memory efficiency rather than a guaranteed wall-clock speedup. Figure~\ref{fig:acc-token}(a) plots the working context against trajectory length. ReAct's context grows almost linearly, whereas ContextFold and MM-ContextFold curb this growth through the state-switching mechanism, at the cost of an initially faster rise than ReAct caused by the longer structured prompts that folding requires. By confining raw images to branch contexts, MM-ContextFold further slows the growth at later steps. AgentFold maintains the smallest context throughout, but its accuracy drops markedly as trajectories lengthen (Figure~\ref{fig:acc-token}(b)). Although accuracy declines on longer trajectories for all methods, MM-ContextFold consistently remains the most accurate, suggesting that the proposed framework better preserves task-relevant information over extended interactions.

\subsection{Case Study}
% To examine how visual complexity affects each method, we conduct a case study on MMSearch-Plus, which features queries with varying numbers of query images. Table~\ref{tab:case-mmsplus} groups samples by image count and reports accuracy alongside tool calls (\textit{Visual Search} / total). All methods degrade as image count increases, with MM-ContextFold retaining the highest accuracy in all three groups (38.1\%, 35.6\%, 27.6\%). Meanwhile ReAct's \textit{Visual Search} usage stays nearly flat, whereas the other methods scale up to 2.2--2.7 calls, suggesting that it underutilizes visual tools on multi-image input. AgentFold yields the lowest accuracy in all three groups (24.9\%, 21.4\%, 20.9\%), indicating that its multi-scale compression discards fine-grained visual evidence that single- and multi-image queries alike depend on. ContextFold trails ReAct slightly on single-image queries (30.2\% vs.\ 31.8\%) but declines more sharply on multi-image ones (24.1\%, 22.8\%), with total tool calls rising from 5.3 to 16.3, a pattern consistent with the view that without visual scoping, globally visible images make later retrieval less targeted. In contrast, MM-ContextFold allocates images to branch contexts on demand, letting each branch focus on its subtask and fold findings back into the text-only main context, sustaining the highest accuracy at moderate tool-call cost.

To examine how visual complexity affects each method, we group MMSearch-Plus samples by query-image count (Table~\ref{tab:case-mmsplus}) and report accuracy and tool calls (\textit{Visual Search} / total). All methods degrade as image count increases, with MM-ContextFold retaining the highest accuracy in all three groups (38.1\%, 35.6\%, 27.6\%). Meanwhile ReAct's \textit{Visual Search} usage stays nearly flat, whereas the other methods scale up to 2.2--2.7 calls, suggesting that it underutilizes visual tools on multi-image input. AgentFold yields the lowest accuracy in all three groups, indicating that its multi-scale compression discards fine-grained visual evidence that single- and multi-image queries alike depend on. ContextFold trails ReAct slightly on single-image queries (30.2\% vs.\ 31.8\%) but declines more sharply on multi-image ones (24.1\%, 22.8\%), with total tool calls rising from 5.3 to 16.3, consistent with the view that without visual scoping, globally visible images make later retrieval less targeted. In contrast, MM-ContextFold allocates images to branch contexts on demand, letting each branch focus on its subtask and fold findings back into the text-only main context, sustaining the highest accuracy at moderate tool-call cost.

\section{Conclusion}
This paper investigates the role of raw images in multimodal agentic retrieval tasks. A systematic empirical study of approximately 10,000 trajectories reveals that the agent's reliance on raw images is concentrated in a limited number of visual-grounding steps, whose duration varies with task complexity, and that retaining images after their visual cues have been textualized brings little downstream benefit while increasing output entropy. Motivated by these findings, we propose MM-ContextFold, a training-free framework that maintains a persistent, text-only main context for high-level planning and confines raw images to ephemeral branch contexts that load them on demand. Experiments on seven MAR benchmarks across five backbone models show that MM-ContextFold improves average accuracy by 6.3 percentage points over the ReAct baseline while reducing working context length by 27.5\%, with the largest gains on tasks that demand multi-step cross-modal evidence synthesis. In future work, we plan to explore adaptive mechanisms that detect when visual grounding is complete, and to extend the folding principle to other token-intensive modalities such as video.

% \begin{acks}
% This work was supported in part by the National Natural Science Foundation of China, No.: 62476071.
% \end{acks}

\bibliographystyle{ACM-Reference-Format}
\balance   % Balances the two columns on the last page (Sheridan requirement).
\bibliography{sample-base}

\clearpage
\nobalance
\appendix
\raggedbottom
\setlength{\emergencystretch}{1.5em}

\section{Dataset}
\label{app:benchmarks}
This section provides details on the seven benchmarks used in this paper, complementing the summary in \S\ref{subsec:exp-setup}. We first present the benchmark taxonomy and category assignment rationale (Table~\ref{tab:appendix-benchmarks}), then describe the unified task interface and report benchmark sizes (Table~\ref{tab:appendix-dataset-size}), and finally document the data source and sampling policy for each dataset (Table~\ref{tab:appendix-data-source}).

\begin{table*}[t]
\caption{Benchmark taxonomy used in this work. We summarize the aspects of each benchmark that are most relevant to multimodal agentic retrieval: how the query image contributes, how much external retrieval is required, and why the benchmark belongs to a particular category.}
\centering
\small
\renewcommand{\arraystretch}{1.25}
\setlength{\tabcolsep}{5pt}
\begin{tabularx}{\textwidth}{l l l X X}
\toprule
\textbf{Benchmark} & \textbf{Category} & \textbf{Image Regime} & \textbf{Role of the Query Images} & \textbf{External Retrieval} \\
\midrule
SimpleVQA & Atomic factuality & Single image & Identifies an entity, artifact, landmark, or product whose name must be recognized. & Answer depends on encyclopedic or attribute knowledge not reliably stored in parametric memory. \\
\addlinespace[2pt]
WorldVQA & Atomic factuality & Single image & Provides the entity to be identified, often involving long-tail objects, places, or culturally specific items. & Retrieval is needed to ground rare or tail knowledge once the visual referent is identified. \\
\midrule
MMSearch & Dynamic info-seeking & Single image & Anchors a current event, organization, scene, or object that must first be identified. & Answer depends on fresh or specialized web information, making static memorization insufficient. \\
\addlinespace[2pt]
LiveVQA & Dynamic info-seeking & Single image & Serves as an entry point to a time-sensitive or domain-specific query. & Required evidence often comes from recent webpages or niche sources that must be actively verified. \\
\midrule
BC-VL & Visual deep research & Single image & Contains a key clue, but the referent is visually ambiguous, indirect, or difficult to name from a single glance. & Requires chained retrieval and cross-source synthesis rather than one-shot identification. \\
\addlinespace[2pt]
MMS-Plus & Visual deep research & Single/multi-image & In multi-image queries, relevant clues are distributed across images; the agent must determine which image matters for which subtask. & Needs to combine evidence across images and sources, not just identify each clue. \\
\addlinespace[2pt]
VDR-Bench & Visual deep research & Fine-grained cue & Contributes subtle evidence such as text fragments, layout cues, or object-level details supporting a longer reasoning chain. & Final answer emerges only after multi-step retrieval, verification, and synthesis over heterogeneous evidence. \\
\bottomrule
\end{tabularx}
\label{tab:appendix-benchmarks}
\end{table*}

\begin{table}[t]
\caption{Reported benchmark sizes and image statistics. The reported set is either the released benchmark split or a stratified subset from the public pool. For MMS-Plus, statistics are over the full normalized release.}
\centering
\small
\renewcommand{\arraystretch}{1.15}
\setlength{\tabcolsep}{6pt}
\begin{tabular}{@{}l r r r@{}}
\toprule
\textbf{Benchmark} & \textbf{\#Samples} & \textbf{Avg.\,\#Imgs} & \textbf{Max\,\#Imgs} \\
\midrule
SimpleVQA  & 300 & 1.00 & 1 \\
WorldVQA   & 200 & 1.00 & 1 \\
MMSearch   & 171 & 1.00 & 1 \\
LiveVQA    & 245 & 1.00 & 1 \\
BC-VL      & 300 & 1.00 & 1 \\
MMS-Plus   & 311 & 1.42 & 5 \\
VDR-Bench  & 200 & 1.00 & 1 \\
\midrule
\textbf{Total} & \textbf{1,727} & -- & -- \\
\bottomrule
\end{tabular}
\label{tab:appendix-dataset-size}
\end{table}

\begin{table*}[t]
\caption{Data source and sampling policy. ``Pool'' denotes the released source set available after normalization. ``Reported'' denotes the set size used in this paper.}
\vspace{-0.2cm}
\centering
\small
\renewcommand{\arraystretch}{1.25}
\setlength{\tabcolsep}{5pt}
\begin{tabularx}{\textwidth}{@{}l l r r X@{}}
\toprule
\textbf{Benchmark} & \textbf{Source} & \textbf{Pool} & \textbf{Reported} & \textbf{Sampling Policy} \\
\midrule
SimpleVQA & \cite{cheng2025simplevqa}, WebWatcher~\cite{geng2026webwatcher} & 300 & 300 & Used directly without additional sampling. \\
\addlinespace[2pt]
MMSearch & \cite{jiang2025mmsearch}, WebWatcher~\cite{geng2026webwatcher} & 171 & 171 & Used directly without additional sampling. \\
\addlinespace[2pt]
LiveVQA & \cite{fu2025seeking}, WebWatcher~\cite{geng2026webwatcher} & 245 & 245 & Used directly without additional sampling. \\
\addlinespace[2pt]
MMS-Plus & \cite{tao2026mmsearch} & 311 & 311 & Used directly after normalization. Image statistics computed over the full release preserving category proportions. \\
\midrule
BC-VL & WebWatcher~\cite{geng2026webwatcher} & 400 & 300 & All 200 Level-2 examples adopted directly; 100 Level-1 examples sampled from the 200 Level-1 pool. \\
\addlinespace[2pt]
WorldVQA & \cite{zhou2026worldvqa} & 3,000 & 200 & Stratified subset by us. Proportional sampling across the 8 non-People categories preserving original ratios. \\
\addlinespace[2pt]
VDR-Bench & \cite{zeng2026vision} & 2,000 & 200 & Stratified subset by us. Preserves released difficulty buckets and topical categories proportionally. \\
\bottomrule
\end{tabularx}
\label{tab:appendix-data-source}
\end{table*}

\subsection{Category Assignment Rationale}
As summarized in Table~\ref{tab:appendix-benchmarks}, our three-way grouping is motivated by the role that visual grounding plays in the subsequent retrieval trajectory.
\emph{Atomic factuality} tasks usually require a relatively short chain: once the image helps identify the referent, the remaining work is largely factual lookup.
\emph{Dynamic information-seeking} tasks additionally require fresh or niche web evidence, but the image still mainly acts as an anchor for retrieving the correct entity or event.
In contrast, \emph{Visual deep research} tasks require multiple rounds of disambiguation, verification, and evidence composition, and therefore place a heavier burden on both visual grounding and long-horizon context management.

\subsection{Unified Task Interface}
All seven benchmarks are normalized into a common MAR interface consisting of a natural-language question $q$, an ordered list of query images $\mathcal{I}$, and a gold answer $a^{*}$.
If a benchmark provides answer options, the options are appended to the question text; otherwise the agent produces a free-form short answer.
This normalization enables all methods to share the same tool interface, answer slot, stopping rule, and evaluation protocol.

As shown in Table~\ref{tab:appendix-dataset-size}, the reported benchmark collection contains 1,727 samples in total under this reporting protocol.
Among the seven benchmarks, MMSearch-Plus is the only one with frequent
multi-image queries. In the full normalized MMSearch-Plus release of 311 examples, 222 are single-image questions and 89 are multi-image questions (54 with two images, 30 with three, 4 with four, and 1 with five), yielding 1.42 query images per sample on average.

\subsection{Data Source and Sampling Policy}

Table~\ref{tab:appendix-data-source} summarizes how each reported set is obtained. The key distinction is between (i) datasets for which we directly adopt the released benchmark split used in prior multimodal-agent evaluations, and (ii) datasets for which we construct a smaller evaluation subset from a larger released pool using dataset-specific sampling policies based on the available difficulty/category metadata.

\textbf{Comparison with WebWatcher benchmark usage.}
Among the benchmarks in our paper, MMSearch, LiveVQA, SimpleVQA, and
BC-VL overlap with the benchmark interface released by WebWatcher
\cite{geng2026webwatcher}. For MMSearch, LiveVQA, and SimpleVQA, we follow the released evaluation split directly. For BrowseComp-VL, we construct a 300-example evaluation set from the released 400-example WebWatcher pool by retaining all 200 Level-2 examples and sampling 100 examples from the 200 Level-1 pool. By contrast, WorldVQA and VDR-Bench are not part of the original WebWatcher benchmark suite, and our 200-example evaluation sets are new proportional subsets constructed from the larger public WorldVQA and VDR-Bench pools.

\textbf{Sampling strategy for constructed subsets.}
For datasets where we construct a smaller evaluation subset, we do not sample uniformly at random. Instead, we preserve the benchmark structure exposed by the released metadata.
For BrowseComp-VL, we retain all 200 Level-2 examples and sample 100 of the 200 Level-1 examples, so that both difficulty levels remain represented and the harder level is kept in full while the evaluation size stays manageable.
For WorldVQA, this means proportional sampling over the eight non-People
categories used in our evaluation. For VDR-Bench, this means preserving both difficulty buckets (e.g., \texttt{base}, \texttt{hard}, \texttt{hard1}) and topical categories encoded in the public example IDs. This design reduces the risk that the sampled subset over-represents only easy categories, only hard categories, or otherwise distorts the released benchmark structure.

\textbf{Data release.}
All normalized benchmark files used in this paper, including the directly
adopted splits, the sampled evaluation subsets, and the corresponding sample IDs, are released in the public repository referenced in the main paper.

\section{Implementation Details}
\label{app:implementation}

This section supplements the implementation summary in \S\ref{subsec:exp-setup} with reproducibility-relevant details omitted from the main text for space.

\subsection{Infrastructure and Hyperparameters}

\begin{table}[t]
\caption{Backbone models, serving configuration, and shared hyperparameters.}
\vspace{-0.2cm}
\centering
\small
\renewcommand{\arraystretch}{1.15}
\setlength{\tabcolsep}{5pt}
\begin{tabular}{@{}l l@{}}
\toprule
\multicolumn{2}{@{}l}{\textbf{Backbone models}} \\
\midrule
Gemini-3-Flash   & OpenRouter API \\
GPT-5.2          & OpenRouter API \\
Qwen3.5-9B       & OpenRouter API \\
Qwen3.5-27B      & OpenRouter API \\
Qwen3.5-35B-A3B  & OpenRouter API \\
\midrule
\multicolumn{2}{@{}l}{\textbf{Shared hyperparameters}} \\
\midrule
Temperature / Top-$p$            & 0.2\,/\,0.95 \\
$T_{\max}$ (total LLM calls)     & 20 (all methods) \\
$B_{\max}$ (max branch steps)    & 5 (MM-ContextFold/ContextFold) \\
Top-$k$ retrieval results               & 10 (\textit{Text Retrieval/Visual Search}) \\
Independent trials                & 3 \\
\bottomrule
\end{tabular}
\label{tab:impl-config}
\end{table}

Table~\ref{tab:impl-config} lists the five backbone models used in the main benchmark evaluations, which are accessed through the OpenRouter API. The local deployment used for the entropy probe and paired image-removal analyses is described below. We use \texttt{temperature}$\,{=}\,0.2$ and \texttt{top\_p}$\,{=}\,0.95$ for all backbones; no system-level sampling changes are made across methods or benchmarks.

\textbf{Local deployment for entropy analysis.}
The entropy probe (\S\ref{subsec:pilot-entropy}) requires token-level log-probabilities unavailable from proprietary APIs.
We therefore deploy the three Qwen3.5 backbones locally on 8$\times$NVIDIA H20 GPUs (96\,GB each) using SGLang with 4-way tensor parallelism for every model.
All local models are served with the same sampling parameters as the OpenRouter runs (temperature$\,{=}\,0.2$, top-$p\,{=}\,0.95$) and return top-20 log-probabilities per token for entropy computation. Each of the three Qwen3.5 backbones is run on the full 1,727-sample benchmark suite; after excluding samples where the agent produced no valid trajectory (e.g., immediate parsing failure), 5,087 trajectories remain and are used in the entropy analysis, as detailed in Table~\ref{tab:entropy-trajectories}.

\begin{table}[t]
\caption{Trajectory counts for the local entropy analysis. Each backbone is run on all 1,727 samples; invalid trajectories are excluded.}
\vspace{-0.2cm}
\centering
\small
\renewcommand{\arraystretch}{1.15}
\setlength{\tabcolsep}{6pt}
\begin{tabular}{@{}l r r r@{}}
\toprule
\textbf{Backbone} & \textbf{Total} & \textbf{Excluded} & \textbf{Valid} \\
\midrule
Qwen3.5-9B      & 1,727 & 45 & 1,682 \\
Qwen3.5-27B     & 1,727 & 21 & 1,706 \\
Qwen3.5-35B-A3B & 1,727 & 28 & 1,699 \\
\midrule
\textbf{Total}   & 5,181 & 94 & \textbf{5,087} \\
\bottomrule
\end{tabular}
\label{tab:entropy-trajectories}
\end{table}

\subsection{Tool-Call Formats and Baseline Adaptation}
\label{subsec:app-tool-format}

Although all methods share the same three retrieval tools and the same total call budget, their action grammars differ because the baselines use different prompt and parsing protocols. To ensure a fair comparison, we preserve each method's native tool-call format and parsing logic while unifying the underlying tool implementations. Table~\ref{tab:tool-call-formats} summarizes the key differences.

\begin{table}[t]
\caption{Tool-call format comparison. Each method retains its action grammar; all are routed to the same backend tool implementations.}
\centering
\small
\renewcommand{\arraystretch}{1.2}
\setlength{\tabcolsep}{4pt}
\begin{tabularx}{\columnwidth}{@{}l X@{}}
\toprule
\textbf{Method} & \textbf{Tool-call format} \\
\midrule
ReAct \newline (WebWatcher) &
JSON inside \texttt{<tool\_call>} tags. \newline
Parameter key: \texttt{"arguments"}. \newline
\texttt{\{"name":"search","arguments": \newline \{"queries":["..."]\}\}} \\
\addlinespace[3pt]
AgentFold &
Same JSON \texttt{<tool\_call>} syntax as ReAct. \newline
Parameter key: \texttt{"arguments"}. Additionally uses \newline
\texttt{<compress>} and \texttt{<motivation>} blocks for multi-scale memory folding. \\
\addlinespace[3pt]
ContextFold &
XML-style tags: \texttt{<function=\textit{name}>} with nested \texttt{<parameter=\textit{key}>value</parameter>}. \newline
Parsed into a flat string-valued argument dict. \\
\addlinespace[3pt]
MM-Context- \newline Fold &
Main state: \texttt{<subtask>} containing a JSON block with \texttt{description}, \texttt{prompt}, and \texttt{assigned\_\allowbreak images}. Branch state: \newline
JSON inside \texttt{<tool\_call>} tags. \newline
Parameter key: \texttt{"parameters"}. \\
\bottomrule
\end{tabularx}
\label{tab:tool-call-formats}
\end{table}

\vspace{+0.2cm}
\noindent\fbox{\parbox{0.96\columnwidth}{\small%
\textbf{ReAct / AgentFold} (JSON): \\[1pt]
\texttt{<tool\_call>} \\
\texttt{\ \{"name":"lens\_search",} \\
\texttt{\ \ "arguments":\{"image\_id":"IMG\_001"\}\}}
\texttt{</tool\_call>} \\[4pt]
\textbf{ContextFold} (XML-style): \\[1pt]
\texttt{<function=lens\_search>} \\
\texttt{\ <parameter=image\_id>IMG\_001</parameter>}
\texttt{</function>}
}}
\vspace{+0.2cm}

\textbf{Preserving action grammars.} Following the adaptation policy in \S\ref{subsec:exp-setup}, we preserve each baseline's context-management logic and action grammar while exposing the shared multimodal inputs and tool interfaces. Parsed tool names and arguments are routed to the same backend APIs (Serper, Jina Reader, and Google Lens). This controls tool access while retaining the method-specific prompts and parsers. As a concrete example, invoking \textit{Visual Search} on image \texttt{IMG\_001} is expressed differently across methods, yet both are routed to the same Google Lens API call.

\textbf{Our format choice.}
For MM-ContextFold, we adopt the JSON-based \texttt{<tool\_call>} format that has become the \textit{de facto} standard in recent agentic retrieval systems~\cite{yao2023react,chu2026redsearcher,geng2026webwatcher,li2025websailor}.
Compared with ContextFold's XML-style tags, JSON tool calls are natively supported by most LLM serving frameworks and align with the function-calling conventions of mainstream model providers, reducing parsing overhead and backbone-specific adaptation.

\subsection{Branch Execution and Folding}

This subsection describes implementation-level details that go beyond the formal definitions in \S\ref{subsec:unified-interface}--\ref{subsec:phase2-grounding}.

\textbf{Tool-use pattern within a branch.}
When a regular retrieval branch receives non-empty \texttt{assigned\_\allowbreak images}, it begins with \texttt{lens\_search} for visual grounding, then typically shifts to \texttt{search} and \texttt{visit} once a candidate entity or clue is grounded. Each tool call carries a goal field specifying the retrieval intent, reducing redundant calls. This staged pattern aligns with the empirical finding in Figures~\ref{fig:intro} and~\ref{fig:pilot}: raw images are most useful at the beginning of a subtask, while later steps are text-dominant.

\textbf{Folding boundary.} The \texttt{RETURN} payload is plain text summarizing the current hypothesis, supporting evidence, and remaining uncertainty. Of the branch-local contents, only this return message is written to the persistent main context; the full branch-local multimodal trajectory is discarded. The main-state reasoning and the corresponding \texttt{BRANCH} action are also retained, as specified in Eq.~(\ref{eq:main-update}). If the branch budget $B_{\max}$ is exhausted before an explicit \texttt{RETURN}, the final call is constrained to summarize partial findings and terminate.

\begin{figure}[t]
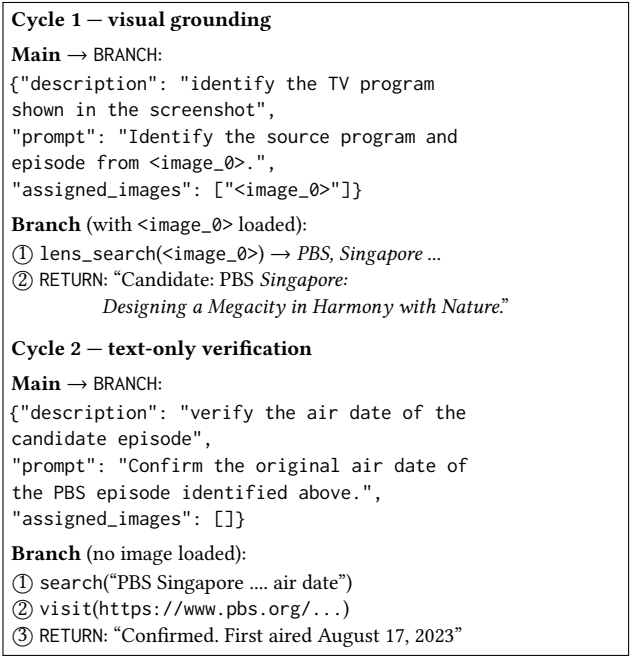

\centering
\small
\fbox{\parbox{0.95\columnwidth}{%
\textbf{Cycle 1 --- visual grounding} \\[3pt]
\textbf{Main} $\rightarrow$ \texttt{BRANCH}: \\[1pt]
\texttt{\{"description": "identify the TV program} \\
\texttt{\ shown in the screenshot",} \\
\texttt{\ "prompt": "Identify the source program and} \\
\texttt{\ episode from <image\_0>.",} \\
\texttt{\ "assigned\_images": ["<image\_0>"]\}} \\[3pt]
\textbf{Branch} (with \texttt{<image\_0>} loaded): \\[1pt]
\textcircled{\small 1} \texttt{lens\_search}(\texttt{<image\_0>}) $\rightarrow$ \textit{PBS, Singapore ...} \\
\textcircled{\small 2} \texttt{RETURN}: ``Candidate: PBS \textit{Singapore:} \\
\hspace*{1.2cm}\textit{Designing a Megacity in Harmony with Nature}.'' \\[6pt]
\textbf{Cycle 2 --- text-only verification} \\[3pt]
\textbf{Main} $\rightarrow$ \texttt{BRANCH}: \\[1pt]
\texttt{\{"description": "verify the air date of the} \\
\texttt{\ candidate episode",} \\
\texttt{\ "prompt": "Confirm the original air date of} \\
\texttt{\ the PBS episode identified above.",} \\
\texttt{\ "assigned\_images": []\}} \\[3pt]
\textbf{Branch} (no image loaded): \\[1pt]
\textcircled{\small 1} \texttt{search}(``PBS Singapore .... air date'') \\
\textcircled{\small 2} \texttt{visit}(\texttt{https://www.pbs.org/...}) \\
\textcircled{\small 3} \texttt{RETURN}: ``Confirmed. First aired August 17, 2023''
}}
\caption{Worked example showing two consecutive branch cycles: the first loads a query image for visual grounding; the second assigns no image and performs text-only verification.}
\label{fig:worked-example}
\end{figure}

\textbf{Error handling.} Malformed model outputs (e.g., missing action fields, batched tool calls) trigger a correction re-prompt that feeds the malformed output back to the model with an error message. The re-prompt counts toward the branch budget $B_{\max}$ and the global budget $T_{\max}$. Figure~\ref{fig:worked-example} illustrates two consecutive branch cycles on the same query after initialization, contrasting a visual-grounding branch ($\mathcal{A}_t \neq \varnothing$) with a text-only verification branch ($\mathcal{A}_t = \varnothing$). Algorithm~\ref{alg:mmcontextfold} summarizes the overall inference-time control flow.

\begin{algorithm}[t]
\caption{Inference-time control flow of MM-ContextFold.}
\label{alg:mmcontextfold}
\small
\begin{algorithmic}[1]
\STATE Assign each query image $I_k$ a stable identifier $\alpha_k$, build the registry $\mathcal{P}=\{(\alpha_k, I_k)\}$, and initialize the main context $\tau_1=(q,\mathcal{A})$ with $\mathcal{A}=\{\alpha_k\}$.
\STATE Initialization cycle: issue \texttt{BRANCH}$(s_{\mathrm{init}}, \mathcal{A})$, which loads all query images and returns $e_{\mathrm{init}}=\{d_k\}$, one textual description per image; fold $e_{\mathrm{init}}$ into the main context.
\WHILE{answer not finalized and global budget $T_{\max}$ not exhausted}
\STATE Main state emits \texttt{FINISH}$(a)$ or \texttt{BRANCH}$(s_t, \mathcal{A}_t)$.
\IF{\texttt{FINISH}$(a)$}
\STATE Terminate and output $a$.
\ELSE
\STATE Create branch: inherit the text-only main context, load only the images in $\mathcal{A}_t$.
\STATE Roll out the branch for at most $B_{\max}$ steps; each step emits a tool call or \texttt{RETURN}$(e_t)$.
\STATE Fold $e_t$ into the main context; discard the branch trajectory.
\ENDIF
\ENDWHILE
\end{algorithmic}
\end{algorithm}

\subsection{Empirical Analysis Protocol}
\label{subsec:app-analysis-protocol}

The methodology of the three pilot analyses (grounding annotation, entropy probe, and image-removal ablation) is described in \S\ref{subsec:pilot-surface-deep}--\S\ref{subsec:pilot-acc}.
Below we document implementation-level details not covered in the main text.

\textbf{Grounding annotation pipeline.}
The annotator is Gemini-3-Flash, prompted with the original question, the query images, and the trajectory steps in chronological order; each step $r_t$ is labeled with respect to the text of the preceding steps, following the counterfactual rule stated in the prompt.
The three mutually exclusive categories used in \S\ref{subsec:pilot-surface-deep} are obtained from two binary labels produced in two passes: \texttt{image\_\allowbreak needed} (whether the step extracts new visual information directly from the raw images) and \texttt{image\_\allowbreak info\_\allowbreak used} (whether the step uses any image-derived information at all). \textit{Image Needed} corresponds to \texttt{image\_\allowbreak needed}=true, \textit{Image-Derived} to \texttt{image\_\allowbreak needed}=false with \texttt{image\_\allowbreak info\_\allowbreak used}=true, and \textit{Text Only} to both labels being false.
Three human annotators independently labeled a random subsample of 500 steps; the LLM annotations achieve 93.5\% agreement with the majority human vote.
The annotation prompts are provided in Appendix~\ref{app:prompts}.

\textbf{Entropy computation details.}
The full-vocabulary entropy $\mathbf{H}(\tau_t)$ defined in \S\ref{subsec:pilot-entropy} is estimated from the top-20 log-probabilities returned per token: the residual probability mass $p_{\mathrm{rest}} = 1 - \sum_{i=1}^{20} p_i$ is treated as a single tail bin contributing $-p_{\mathrm{rest}} \log p_{\mathrm{rest}}$, so the reported values are a grouped (lower-bound) estimate of the exact vocabulary-level entropy, computed identically for $\tau_t$ and $\tau_t^{\text{IR}}$. For thinking-enabled Qwen3.5 backbones, we compute entropy only over tokens inside the \texttt{<think>}$\ldots$\texttt{</think>} reasoning segment, excluding the final answer tokens; this ensures that $\Delta H_t$ reflects changes in the model's internal reasoning distribution rather than answer formatting.

\subsection{Scope of Visual Analysis}
Our study focuses on the retention of \emph{original query images}.
Unlike retrieved images that enter the context at varying points depending on when the agent invokes Visual Search, query images are present from step~1 and shared uniformly across all samples, making them the cleanest unit for isolating a context-management question: once a visual clue has been extracted and verbalized, should the raw input image still remain globally visible?

In our ReAct-based analysis, retrieved images from \textit{Visual Search} are excluded as the primary analysis unit for three reasons.
First, their content is non-deterministic and search-engine dependent: the same query image may yield different retrieved results across runs due to ranking changes, freshness, and index updates, making controlled before-and-after comparisons unreliable.
Second, analyzing retrieved-image retention would conflate two factors we wish to keep separate: \emph{retrieval quality} (whether the search engine returns relevant images) and \emph{context-retention policy} (whether keeping those images in subsequent steps helps or hurts reasoning).
Third, retrieved images enter the ReAct context at varying trajectory positions depending on when the agent invokes Visual Search, whereas query images are present from step~1 onward and uniformly shared across all samples. This uniformity is essential for the step-level aggregation in Figures~\ref{fig:intro} and~\ref{fig:pilot}.
By focusing on query-image retention, we obtain a cleaner intervention with a fixed, sample-independent baseline for studying when persistent raw visual context helps and when it becomes redundant.

\subsection{Evaluation Protocol}

The evaluation pipeline is shared across all methods and backbones; we describe here the implementation details beyond the summary in \S\ref{subsec:exp-setup}.

\textbf{Judge model and prompt.}
We use Gemini-3-Flash (accessed via OpenRouter) as the judge model for all evaluations.
The judge prompt follows the WebWatcher and Qwen-DeepResearch official evaluation template~\cite{geng2026webwatcher, team2025tongyi}, which performs a two-step procedure: (1)~\textit{extract} the core factual claim from the model's response, and (2)~\textit{grade} it against the gold answer as one of three classes: \texttt{CORRECT} (semantically equivalent, ignoring casing, phrasing, and minor typos), \texttt{INCORRECT} (contains a factual contradiction), or \texttt{NOT\_\allowbreak ATTEMPTED} (no clear claim is made).
The judge may emit either the textual grades above or the equivalent letter codes \texttt{A}/\texttt{B}/\texttt{C}, which are mapped to the same three classes by a deterministic parser.
The letter codes are mapped as \texttt{A}=\texttt{CORRECT}, \texttt{B}=\texttt{INCORRECT}, and \texttt{C}=\texttt{NOT\_\allowbreak ATTEMPTED}.
The full judge prompt is provided in Appendix~\ref{app:prompts}.

\textbf{Grading rules.}
For numerical answers, values correct to the last significant figure of the gold target are accepted. When the prediction only addresses the specific question asked, extra detail in the gold target is not penalized. If a response contains both correct and incorrect claims, the grade is \texttt{INCORRECT}.

\textbf{Handling failed trajectories.}
If a trajectory terminates with status other than \texttt{success} (e.g., the agent exhausted its budget without producing a final answer) or returns an empty answer, the sample is automatically graded as \texttt{NOT\_\allowbreak ATTEMPTED} without invoking the judge model.

\textbf{Parallelism and reproducibility.}
Judge calls are parallelized with up to 8 concurrent workers.
The judge uses temperature$\,{=}\,0$ to maximize grading consistency.
Table~\ref{tab:main-results} reports mean accuracy and standard deviation over three independent runs; the supplementary HLE-VL table reports the means. Accuracy is the fraction of evaluated samples graded \texttt{CORRECT}; \texttt{INCORRECT} and \texttt{NOT\_\allowbreak ATTEMPTED} samples remain in the denominator.

\section{Comparison With Context Management}
\label{app:comparison}

This section provides a principled comparison of MM-ContextFold with the context-management baselines.
We organize the analysis around the three design principles introduced in \S\ref{sec:method}: (P1)~\textit{decoupled reasoning and grounding}, (P2)~\textit{ephemeral visual context}, and (P3)~\textit{adaptive image allocation}.

\begin{table*}[t]
\caption{Design-principle comparison. Each method is assessed against the three principles (P1--P3) motivating MM-ContextFold. \cmark\ = fully satisfied, \pmark\ = partially satisfied, \xmark\ = not satisfied. All agentic methods share the backbone, tools, and budget described in \S\ref{subsec:exp-setup}.}
\centering
\small
\renewcommand{\arraystretch}{1.3}
\setlength{\tabcolsep}{5pt}
\begin{tabularx}{\textwidth}{@{}l c c c X@{}}
\toprule
\textbf{Method} & \textbf{P1} & \textbf{P2} & \textbf{P3} & \textbf{How it manages visual context} \\
\midrule
ReAct &
\xmark & \xmark & \xmark &
Append-only mixed-modality history. Raw query images remain globally visible throughout the trajectory; planning and grounding share a single interleaved context. \\
AgentFold &
\pmark & \pmark & \xmark &
Multi-scale textual compression. Planning operates over compressed text (partial P1), and raw images are replaced by textual summaries (partial P2). Compression is lossy, so fine-grained visual cues may be lost and cannot be selectively reloaded later. \\
ContextFold &
\pmark & \xmark & \xmark &
Branch-and-return with modality-symmetric inheritance. Branching provides subtask isolation (partial P1), but raw images in the main context propagate into every subsequent branch, and image assignment is implicit rather than per-subtask. \\
MM-ContextFold &
\cmark & \cmark & \cmark &
Text-only main state for planning (P1); raw images confined to ephemeral branches and discarded after return (P2); each \texttt{BRANCH} action explicitly specifies which images to load, including $\varnothing$ for text-only subtasks (P3). \\
\bottomrule
\end{tabularx}
\label{tab:appendix-comparison}
\end{table*}
\subsection{Adaptation Principle}
We adapt all baselines to the same MAR setting under a strict policy: \emph{preserve each method's original context-management logic and action grammar while unifying the external interfaces} (backbone, tool APIs, answer format, and global call budget $T_{\max}=20$).
The shared interfaces control tool access, while method-specific prompts implement the different context-management designs.

\textbf{ReAct (WebWatcher).}
We directly adopt the publicly released WebWatcher agent, which follows the standard ReAct loop. The agent receives the original question and all query images at step~1, and appends all tool observations (including retrieved images) to a single, ever-growing context. No context-management modification is applied; this serves as the prevailing MAR baseline.

\textbf{AgentFold.}
We preserve AgentFold's original multi-scale textual compression logic: after each tool call, the agent decides whether to compress earlier steps into a textual summary. To adapt it to MAR, we expose the same query-image input and Visual Search tool interface as our method while keeping the compressed memory text-only, faithful to the original design. This means visual information can only survive through textualization, a principled design that is inherently lossy when fine-grained visual details are needed later.

\textbf{ContextFold.}
We preserve ContextFold's original branch-and-return controller, inherited-state design, and its native XML-style tool-call format (see \S\ref{subsec:app-tool-format}). The agent can invoke the same Visual Search tool, but several structural differences distinguish it from MM-ContextFold:
\begin{itemize}[leftmargin=*]
\item \textbf{Per-subtask image allocation.} In ContextFold, raw images present in the main state are automatically inherited by all spawned branches regardless of whether the current subtask requires visual grounding. MM-ContextFold instead lets the main state declare $\mathcal{A}_t \subseteq \mathcal{A}$ (possibly $\varnothing$) per branch, so that later verification branches run text-only without carrying irrelevant image tokens.
\item \textbf{Modality-symmetric inheritance.} ContextFold's inherited state is a full copy of the main context including any raw images, making it modality-symmetric. MM-ContextFold enforces modality-\emph{asymmetric} inheritance: only textual artifacts are inherited from the main state, and raw images are injected on demand.
\end{itemize}

\subsection{Principle-Level Analysis}

\textbf{P1: Decoupled reasoning and grounding.}
ReAct interleaves planning, grounding, and retrieval in a single context with no structural separation. AgentFold partially decouples them by compressing earlier observations, but the compression itself mixes visual and textual information. ContextFold provides subtask isolation via branches, but within each branch planning and grounding are still interleaved, and the main state can contain raw images.
MM-ContextFold enforces a separation: the main state performs planning over text-only artifacts, and visual grounding is confined to branches.

\textbf{P2: Ephemeral visual context.}
MM-ContextFold guarantees that no raw image persists in the main context after a branch returns. In ReAct, images accumulate indefinitely. In ContextFold, images inherited from the main state persist across branches. AgentFold achieves partial ephemerality by compressing images into text, which is a one-way, lossy transformation and is not a reversible scoping mechanism.

\textbf{P3: Adaptive image allocation.}
In MM-ContextFold, each subtask explicitly declares which images it needs via the \texttt{assigned\_\allowbreak images} field, including $\mathcal{A}_t = \varnothing$ for purely textual subtasks.
This per-subtask allocation reflects the observation that different stages of a trajectory have different visual demands (Figure~\ref{fig:pilot}(a)): early grounding steps benefit from direct image access, while later verification steps typically do not.
By contrast, ReAct keeps all images always visible, AgentFold replaces them with compressed text, and ContextFold inherits them globally, each handling image lifecycle at a coarser granularity than per-subtask allocation.

\section{More Results}
\label{app:results}

\subsection{Results on HLE-VL}

To further validate MM-ContextFold on a benchmark outside our main evaluation suite, we report results on \textbf{Humanity's Last Exam-Visual (HLE-VL)}~\cite{geng2026webwatcher}, a highly challenging multimodal benchmark that spans biology, chemistry, CS/AI, engineering, humanities, mathematics, physics, and other disciplines.
HLE-VL is a core benchmark in the WebWatcher evaluation~\cite{geng2026webwatcher}; we adopt its released 330 question visual subset and apply the same evaluation protocol as our main experiments (\S\ref{subsec:exp-setup}). Table~\ref{tab:hle-vl} reports accuracy averaged over three independent runs on two representative backbones. MM-ContextFold achieves the highest accuracy on both backbones, outperforming the ReAct baseline by $+$4.1 percentage points on Gemini-3-Flash and $+$4.7 percentage points on Qwen3.5-35B-A3B. These results are consistent with the main findings in Table~\ref{tab:main-results}: MM-ContextFold provides a consistent advantage over context-management baselines, including on benchmarks not included in our primary evaluation suite.

\begin{table}[t]
\caption{Results on HLE-VL (accuracy, \%). Each entry reports the mean over three independent runs. All agentic methods use the same tool set and budget.}
\centering
\resizebox{\columnwidth}{!}{
\begin{tabular}{@{}l ccccc@{}}
\toprule
\textbf{Backbone} & \textbf{Direct} & \textbf{ReAct} & \textbf{AgentFold} & \textbf{ContextFold} & \textbf{Ours} \\
\midrule
Gemini-3-Flash   & 27.4 & 32.7 & 29.8 & 33.0 & \textbf{36.8} \\
Qwen3.5-35B-A3B & 20.3 & 30.0 & 27.4 & 32.0 & \textbf{34.7} \\
\bottomrule
\end{tabular}
}
\label{tab:hle-vl}
\end{table}

\subsection{Sensitivity to Branch Budget \texorpdfstring{$B_{\max}$}{Bmax}}

The branch budget $B_{\max}$ controls the maximum number of steps allowed within a single branch before the agent must return its findings to the main state.
We set $B_{\max} = 5$ in all main experiments (\S\ref{subsec:exp-setup}).
To assess the sensitivity of this choice, we vary $B_{\max} \in \{2, 3, 4, 5, 6\}$ on Qwen3.5-35B-A3B. For efficiency, we randomly sample 100 examples from each of the seven benchmarks (700 samples in total) and report the average accuracy over this subset; the absolute values are therefore not directly comparable with the full-benchmark results in Table~\ref{tab:main-results}.

\begin{table}[t]
\caption{Sensitivity of MM-ContextFold to the branch budget $B_{\max}$ (average accuracy (\%) on the 700-sample subset, Qwen3.5-35B-A3B). All other hyperparameters are held fixed.}
\centering
\small
\renewcommand{\arraystretch}{1.15}
\setlength{\tabcolsep}{6pt}
\begin{tabular}{@{}l ccccc@{}}
\toprule
& $B_{\max}{=}2$ & $B_{\max}{=}3$ & $B_{\max}{=}4$ & $B_{\max}{=}5$ & $B_{\max}{=}6$ \\
\midrule
Qwen3.5-35B-A3B & 56.9 & 57.8 & 58.4 & \textbf{58.7} & 58.2 \\
\bottomrule
\end{tabular}
\label{tab:bmax-sensitivity}
\end{table}

As shown in Table~\ref{tab:bmax-sensitivity}, accuracy increases steadily from $B_{\max}{=}2$ to $B_{\max}{=}5$, with the largest jump between 2 and 3 ($+$0.9 percentage points).
Performance peaks at $B_{\max}{=}5$ (58.7\%) and drops slightly at $B_{\max}{=}6$ (58.2\%).
The results suggest that a branch budget of 5 provides a favorable trade-off: shorter branches ($B_{\max} \le 3$) may force premature returns before the agent has gathered sufficient evidence, while longer branches ($B_{\max}{=}6$) offer diminishing returns and may accumulate redundant tool calls within a single subtask.
Overall, accuracy varies within a 1.8 percentage-point range across the five settings, indicating that MM-ContextFold is not highly sensitive to this hyperparameter.

\subsection{Trajectory Examples}

A common MM-ContextFold trajectory follows four stages:
(i)~the initialization cycle produces coarse image descriptions that help the main state name candidate entities, source types, or visible keywords;
(ii)~the first image-grounding branch allocates one or a few relevant images and uses \textit{Visual Search} to identify a source page, entity, or clue that cannot be reliably inferred from text alone;
(iii)~subsequent branches become increasingly text-dominant, using \textit{Text Retrieval} and \textit{Web Visit} to verify the initial hypothesis, collect factual evidence, and reconcile conflicting clues;
(iv)~the main state synthesizes the folded textual evidence and outputs the final answer without carrying raw visual tokens throughout the trajectory.

We present three correctly solved examples from the Qwen3.5-35B-A3B backbone, spanning multiple benchmark categories, to illustrate how the framework adapts to varying visual demands and task complexity.

\subsubsection{Case~1: Identifying a Runway Collection from a Street-Style Photo (LiveVQA)}

\textbf{Question.} \textit{``Which organization's runway collection included the ensemble worn by the woman shown walking on the sidewalk, considering the creative director's debut was marked by another label that season?''}
\textbf{Reference answer:} \textit{Shushu/Tong.}
The query image shows a woman on a sidewalk wearing a navy-blue two-piece ensemble with a white-collar crop top and a matching midi skirt with white angular detail.

\begin{itemize}[nosep,leftmargin=1.5em]
\item \textbf{Branch~1} (image: \texttt{<image\_0>}).
Initialization: produces a planner-friendly description of the image, noting the outfit details and street-style setting.

\item \textbf{Branch~2} (image: \texttt{<image\_0>}).
Subtask: \textit{``Identify the woman and the specific fashion outfit. Find which designer/brand this ensemble is from.''}
The branch performs a reverse-image search, identifies the outfit as Shushu/Tong, and returns the brand name with a candidate collection context.

\item \textbf{Branch~3} (image: \texttt{<image\_0>}).
Subtask: \textit{``Confirm the outfit details. Find information about Shushu/Tong's creative director and their debut season.''}
The branch corroborates the brand attribution via additional image-based search and surfaces the designers' names (Liushu Lei and Yutong Jiang).

\item \textbf{Branch~4} (text only).
Subtask: \textit{``Search for Shushu/Tong's creative directors and whether they had a debut at another label that season.''}
With visual grounding complete, this text-only branch searches for the designers' career history and confirms the creative-director context referenced in the question.
\end{itemize}

\textbf{Output:} \textbf{Shushu/Tong} (correct).
The transition from image-assigned branches (1--3) to a text-only branch (4) reflects P3: once the brand identity is visually confirmed, the final verification subtask proceeds without pixel-level access.
The branches address identification, visual verification, and textual fact-checking.

\subsubsection{Case~2: Tracing a University Facility Closure from a Logo (BrowseComp-VL)}

\textbf{Question.} \textit{``Which location of this university in the image experienced the earliest permanent closure following a reduction in operating hours?''}
\textbf{Reference answer:} \textit{Saxbys coffee shop.}
The query image shows the University of Pennsylvania logo and branding.

\begin{itemize}[nosep,leftmargin=1.5em]
\item \textbf{Branch~1} (image: \texttt{<image\_0>}).
Initialization: identifies the image as University of Pennsylvania branding.

\item \textbf{Branch~2} (image: \texttt{<image\_0>}).
Subtask: \textit{``Confirm UPenn branding via image search; then search for campus locations that have permanently closed.''}
The branch confirms the university identity and retrieves a list of historical campus relocations, but these are planned relocations, not closures following operating-hour reductions.

\item \textbf{Branch~3} (text only).
Subtask: \textit{``Search for UPenn extension centers, satellite facilities, or specialized locations that closed after reducing operating hours.''}
The branch surfaces several candidate facilities but does not find a definitive match with the ``reduced hours then closed'' criterion.

\item \textbf{Branch~4} (text only).
Subtask: \textit{``Search for any UPenn location that had reduced hours before closing, including medical, continuing-education, or campus retail locations.''}
This branch discovers the Saxbys coffee shop at 40th and Locust Streets, which eliminated weekend operations in April~2025 and permanently closed on May~2, 2025, after Penn Dental Medicine reclaimed the space.
\end{itemize}

\textbf{Output:} \textbf{Saxbys coffee shop} (correct).
Here the visual-grounding phase is short (a single branch confirms the university identity), after which two text-only branches progressively narrow the search scope (campus relocations $\rightarrow$ satellite closures $\rightarrow$ retail closures with hour reductions), illustrating how the main state uses return summaries to refine subsequent subtask prompts.

\subsubsection{Case~3: Identifying a Butterfly--Plant Ecological Relationship (VDR-Bench)}

\textbf{Question.} \textit{``Which butterfly species in the genus Eurema uses the plant as a host, and what is the ecological significance of its nectar for pollinators?''}
\textbf{Reference answer:} \textit{Eurema mexicana (Mexican Yellow); its nectar supports a wide range of pollinators.}
The query image shows a plant with fern-like compound leaves and small green buds, alongside a butterfly with iridescent green wings and brown edges.

\begin{itemize}[nosep,leftmargin=1.5em]
\item \textbf{Branch~1} (image: \texttt{<image\_0>}).
Initialization: describes both organisms---a plant with fern-like compound leaves and a butterfly with iridescent green wings.

\item \textbf{Branch~2} (image: \texttt{<image\_0>}).
Subtask: \textit{``Identify the plant and butterfly species via image search; then search for which Eurema species uses this plant as a host.''}
The branch identifies the plant as \textit{Acacia angustissima} (Prairie Acacia) and confirms the genus \textit{Eurema} as a candidate host, but the specific species requires further verification.

\item \textbf{Branch~3} (text only).
Subtask: \textit{``Search for which Eurema species use Acacia angustissima as a host plant; find the ecological significance of Acacia nectar for pollinators.''}
No further image access is needed. The branch retrieves documentation confirming \textit{Eurema mexicana} (Mexican Yellow) as the host-associated species and that Acacia nectar supports a wide range of pollinators.
\end{itemize}

\textbf{Output:} \textbf{Eurema mexicana} (correct).
When the question is relatively focused, MM-ContextFold naturally converges with fewer branches, only one visual branch and one text-only branch beyond initialization.
Compared with the four-branch trajectories in Cases~1 and~2, this example shows that the framework does not impose a fixed number of branches but instead lets the main state decide when sufficient evidence has been gathered.

\subsection{Illustrative Image-Dependency Cases}
We also manually checked a small number of examples from the step-level annotation results used for Figure~\ref{fig:pilot}(a) and retained only cases whose original ReAct answers are correct in the corresponding Gemini-3-Flash run.
The annotations referenced below are drawn from the two-pass protocol described in \S\ref{subsec:app-analysis-protocol} and Appendix~\ref{app:prompts}: the first pass determines \texttt{image\_\allowbreak needed} (i.e., whether the step extracts new visual information), and the second pass determines \texttt{image\_\allowbreak info\_\allowbreak used} (i.e., whether the step uses any image-derived information at all).
These examples help interpret the difference between \textit{Image Needed}, \textit{Image-Derived}, and \textit{Text Only}.

\textbf{Example 1: \textit{Image Needed} followed by \textit{Image-Derived} (\texttt{bcvl\_\allowbreak level1\_\allowbreak 12}).}
The question asks how many students at the university shown in the image were commended after placing second in a 2017 tug-of-war competition; the inspected run correctly answers \textit{36}.
At Step~1, the agent uses visual evidence from the university gate to identify \textit{Harbin Engineering University}, making the step clearly \textit{Image Needed}.
At Step~2, the agent performs web search using that identified university name, so the step is no longer directly image-grounded but still relies on information extracted from the image; this is \textit{Image-Derived}.
At Step~3, the final synthesis still depends on the university identity recovered from the image, so it remains \textit{Image-Derived} rather than \textit{Text Only}.

\textbf{Example 2: \textit{Image Needed} followed by genuinely \textit{Text Only} reasoning (\texttt{bcvl\_\allowbreak level1\_\allowbreak 93}).}
The question asks for the Hausdorff dimension of the figure described as being constructed by recursively removing the middle third from a line segment; the inspected Gemini-3-Flash run correctly answers \textit{Cantor set, $\log 2/\log 3$}.
The trajectory contains three steps.
At Step~1, the agent calls \texttt{lens\_search} on the query image, which is classified as \textit{Image Needed} because the agent directly submits the image to the visual-search tool.
Notably, the image itself depicts a colorful fractal tree, which does not visually resemble the Cantor set described in the question text.
At Step~2, the agent recognizes this mismatch, where the textual description \textit{``recursively removing the middle third from a line segment''} already identifies the mathematical object as the Cantor set, and calls \texttt{web\_search} with a text-based query grounded entirely in the question text rather than any visual detail.
The annotator classifies this step as \textit{Text Only} (\texttt{image\_\allowbreak info\_\allowbreak used\,=\,false}) because the search query derives from the mathematical description in the question, not from the image content. At Step~3, the agent synthesizes the retrieved mathematical facts and produces the final answer; this step is also \textit{Text Only} for the same reason. This example shows that later non-image steps are not always merely \textit{Image-Derived}: when the decisive evidence is already present in the question text, the trajectory can genuinely switch to a purely textual regime even though the first step used the image.

\subsection{Representative Failure Modes}
\textbf{Overly coarse initialization summaries.}
If the initialization cycle misses a small but decisive clue, the main state may assign an under-specified first branch. This does not invalidate the dual-state design, but it highlights a trade-off: the initialization text should be concise enough for planning while still preserving salient source cues.

\textbf{Search-engine drift and noisy visual retrieval.} Some failures arise when reverse-image retrieval returns near-duplicate distractors, low-quality mirrors, or pages whose titles are only weakly related to the query image. These failures are partly orthogonal to context management: even a better planner cannot recover if the evidence returned by the external search engine is systematically misleading.

\textbf{Webpage parsing noise.}
Long or template-heavy webpages can bury the relevant evidence in surrounding boilerplate. In such cases, the branch return may still contain imperfect or incomplete textual evidence even when the correct source page was found.

\section{Prompts}
\label{app:prompts}

This section reproduces the full prompt templates used in the final reported runs. We present seven prompts in total: the shared system prompt (\S\ref{app:prompts}.1), the three MM-ContextFold state prompts, including main, branch, and initialization (\S\ref{app:prompts}.2--4), the answer-evaluation judge prompt (\S\ref{app:prompts}.5), and the two-stage grounding-analysis annotator prompts used for the pilot study in Figure~\ref{fig:pilot}(a) (\S\ref{app:prompts}.6--7).

\subsection{System Prompt}
All states share a single system prompt that defines the dual-role architecture:

\begin{lstlisting}[style=promptbox]
You are a research agent that answers questions by planning and executing focused subtasks. You operate in two states: MAIN (planner) or BRANCH (executor).
\end{lstlisting}

\subsection{Main-State Prompt}
The main-state prompt presents the agent as a research strategist.
It requires exactly one action per response: either create a branch subtask (\texttt{BRANCH}) or provide the final answer (\texttt{FINISH}).
The \texttt{assigned\_\allowbreak images} field is the mechanism for adaptive image allocation (P3): it specifies which query images should be loaded into the next branch, and may be an empty list for text-only subtasks.
The \texttt{image\_\allowbreak context} slot holds the image descriptions $e_{\mathrm{init}}$ produced by the initialization cycle (\S\ref{subsec:phase1-init}).

\begin{lstlisting}[style=promptbox]
**STATE: MAIN** -- You are the research strategist.

## Workflow
1. Analyze the question and image descriptions
2. Identify what information is still missing
3. Create a focused subtask to gather it
4. Review subtask results, then create another subtask or answer
5. Keep a global round budget in mind; when budget is low, request a synthesis-focused subtask or provide the best calibrated answer.

## Creating a Subtask
Output ONE subtask per response. The executor (BRANCH) will inherit your full context and can use web search, page reading, and reverse image search.

<subtask>
{"description": "3-5 word summary",
 "prompt": "Clear task prompt with objectives. State what to search for, what specific information to extract, and what to return.",
 "assigned_images": ["<image_0>"]}
</subtask>

**`assigned_images` guide:**
- Only assign images that the subtask **directly needs** (e.g., for reverse image search, reading text/numbers, identifying visual details).
- For pure text-based research (e.g., searching a name or fact), use an empty list [] -- no images needed.
- Each image is only loaded inside the BRANCH that receives it, keeping context focused.

## Answering
When evidence is sufficient:
<answer>your final answer</answer>

## Image Context
{image_context}

## Question
{question}
\end{lstlisting}

\paragraph{Continuation prompt.}
After each branch returns, the main state receives the following continuation prompt, which presents the branch findings and requires exactly one follow-up action:

\begin{lstlisting}[style=promptbox]
## Subtask [{description}] completed

{return_message}

---
Review the findings above. Then take **exactly one** action:

**Option A -- Create another subtask:**
<subtask>
{"description": "...", "prompt": "...", "assigned_images": [...]}
</subtask>

**Option B -- Provide your final answer:**
<answer>your final answer</answer>
\end{lstlisting}

\paragraph{Force-answer prompt.}
When the global budget $T_{\max}$ is reached without a final answer, the last main-state call uses the following prompt:

\begin{lstlisting}[style=promptbox]
Maximum research rounds reached. You MUST provide your best answer now based on all evidence gathered.
<answer>your final answer</answer>
\end{lstlisting}

\subsection{Branch-State Prompt}
The branch-state prompt presents the agent as a focused executor that inherits the current main-state text and optionally receives a subset of raw query images.
In addition to the basic execution loop, the prompt includes convergence-focused rules: a per-branch budget of 5 tool calls, an evidence-validation policy requiring at least two independent signals, and anti-loop and source-quality heuristics.

\begin{lstlisting}[style=promptbox]
**STATE: BRANCH** -- Execute the assigned research task.

## Available Tools
<tools>
{"name": "search",
 "description": "Perform a Google web search.",
 "parameters": {"query": "string", "goal": "string"}}
{"name": "visit",
 "description": "Visit a URL to extract detailed content.",
 "parameters": {"url": "string", "goal": "string"}}
{"name": "lens_search",
 "description": "Perform Google Lens reverse image search.",
 "parameters": {"image_id": "string", "goal": "string"}}
{"name": "return",
 "description": "Report findings back to MAIN.",
 "parameters": {"message": "string"}}
</tools>

## Response Format
Think step-by-step, then call **one tool** per response:
<tool_call>
{"name": "tool_name", "parameters": {"key": "value"}}
</tool_call>

## Rules
1. Focus exclusively on the assigned task.
2. When done, you **must** call `return` to report findings back to MAIN.
3. Budget: finish within at most 5 tool calls; if uncertainty remains, call `return` with the best-supported conclusion and explicit uncertainty.
4. Evidence policy: before `return`, try to validate the key claim with at least two independent signals (e.g., lens_search + search, or search + visit).
5. Anti-loop: do not repeat near-duplicate queries more than twice; if results are noisy, pivot strategy, then conclude.
6. Source quality: prefer official or domain-relevant sources over forums/aggregators when evidence conflicts.

## Task
{task_prompt}
\end{lstlisting}

\subsection{Initialization Prompt}
The initialization prompt instantiates the initialization subtask $s_{\mathrm{init}}$ (\S\ref{subsec:phase1-init}). It is intentionally short and task-agnostic, and is applied to each query image loaded in the initialization branch.
Its purpose is to produce a planner-friendly description $d_k$ of each query image, not to answer the question; the resulting descriptions form $e_{\mathrm{init}}$ and are folded into the main context as the image context.

\begin{lstlisting}[style=promptbox]
Briefly describe this image for a research planner who cannot see it.

1. What the image shows (subject, type of content)
2. Any visible text or numbers
3. What type of source it appears to be from

Output as JSON:
{"summary": "...", "source_hint": "..."}
\end{lstlisting}

\subsection{Judge Prompt}
The answer evaluator follows the two-stage extract-then-grade protocol of the WebWatcher and Qwen-DeepResearch evaluation template~\cite{geng2026webwatcher, team2025tongyi}, as described in \S\ref{subsec:exp-setup} and Appendix~\ref{app:implementation}.
The full prompt is reproduced below.

\begin{lstlisting}[style=promptbox]
You are grading a predicted answer against a gold target.

Task -- two steps:
1. EXTRACT: From the predicted answer, extract the core factual claim that directly answers the question. Write it as a short phrase.
2. GRADE: Compare the extracted claim with the gold target and assign exactly one grade.

Grading rules:
- CORRECT: The extracted claim is semantically equivalent to the gold target. Ignore casing, punctuation, phrasing, order, or minor typos. Hedging is acceptable if the correct answer is clearly stated.
- INCORRECT: The extracted claim contains a factual statement that contradicts the gold target (even if hedged). If correct and incorrect claims coexist, grade INCORRECT.
- NOT_ATTEMPTED: No clear factual claim is made (e.g. "I don't know"), or only partial information is given with no contradictions.

Additional rules:
- For numbers, accept values correct to the last significant figure of the gold target.
- The prediction only needs to address what the question asks, even if the gold target contains extra detail.
- Do not penalize omission of info clearly inferable from the question.

Question: {question}
Gold target: {correct_answer}
Predicted answer: {response}

Reply in exactly this format (2 lines, nothing else):
Extracted: <core answer in a short phrase>
Grade: <A, B, or C>
\end{lstlisting}

\subsection{Grounding-Analysis Annotator Prompt}
The grounding annotator (Stage~1) used in Figure~\ref{fig:pilot}(a) receives the question and the step-by-step trajectory in chronological order, then classifies each step with a counterfactual rule: \emph{could this step still be produced if the raw image were removed while all previous textual output were retained?}
The full prompt is reproduced below.

\begin{lstlisting}[style=promptbox]
You are analyzing a multi-step reasoning trajectory from a vision-language agent. The agent was given a question about an image and used tools (web_search, lens_search, visit) to find the answer.

Your task: For each assistant step below, determine whether the step requires direct access to the original image to obtain NEW visual information, or whether it merely reuses image-derived information that was already textualized in a previous step.

## Classification Labels

**"image_needed" = true**: This step extracts NEW information directly from the image that has NOT appeared in any previous step's text. Examples:
- FIRST mention of what the image depicts
- Using lens_search / reverse image search on the original image
- Noticing a NEW visual detail not previously described
- Re-examining the image to verify or correct a previous observation

**"image_needed" = false**: This step does NOT require the original image. All image-related information it uses was already converted to text in earlier steps. Examples:
- Searching for an entity identified from the image in a PREVIOUS step
- Reasoning about facts retrieved from web search results
- Mentioning image content that was already fully described earlier
- The step is purely text-based: processing search results, synthesizing information, or giving a final answer

## Key Principle
Ask yourself: "If I removed the original image from this step's input but kept all previous steps' text output, could this step still produce the same result?" If YES -> image_needed = false.

## Question
{question}

## Steps (in chronological order)
{steps_text}

## Output Format
Return a JSON array with one object per step:
[{"step": 1, "tool": "...", "image_needed": true/false, "reason": "brief explanation (1 sentence)"}]
\end{lstlisting}

\subsection{Image-Info Reuse Annotator Prompt}
Stage~2 further refines each step's label by asking whether the step uses \emph{any} image-derived information, even if that information was already textualized in earlier steps.
Combined with Stage~1, this produces the three-way decomposition used in Figure~\ref{fig:pilot}(a):
\begin{itemize}[nosep,leftmargin=1.5em]
  \item \textit{Image Needed} --- \texttt{image\_needed}=true.
  \item \textit{Image-Derived} --- \texttt{image\_needed}=false, \texttt{image\_info\_used}=true.
  \item \textit{Text Only} --- \texttt{image\_needed}=false, \texttt{image\_info\_used}=false.
\end{itemize}

\begin{lstlisting}[style=promptbox]
Each step has already been labeled with:
  image_needed: whether this step requires DIRECT access to the original image

Your additional task: for each step, determine **image_info_used** -- whether this step's reasoning or action makes use of ANY image-derived information, regardless of whether it comes directly from the original image or was textualized in a previous step.

## Classification for image_info_used

**true**: This step uses image information in any form. Examples:
- The step directly examines the original image (image_needed=true implies this)
- The step's search query or reasoning references something identified from the image in a previous step
- The step's reasoning mentions visual details that came from the image

**false**: This step has NO dependency on image information whatsoever. Examples:
- The step processes purely factual web results unrelated to the image content
- The step reasons about dates, numbers, or facts that could have been in a text-only question
- The step synthesizes information that has no connection to the original image

## Key Principle
Ask yourself: "If this had been a text-only question (no image at all), could this step still appear in an equivalent reasoning trajectory?" If YES -> image_info_used = false.

## Question
{question}

## Steps (in chronological order)
{steps_text}

## Output Format
[{"step": 1, "tool": "...", "image_needed": true/false, "image_info_used": true/false, "reason": "brief explanation (1 sentence)"}]

For steps where image_needed=true, image_info_used must also be true.
\end{lstlisting}

\end{document}